\pdfoutput=1
\documentclass[acmtog]{acmart}
\usepackage{bm}
\usepackage{xspace}

\newcommand{\arxivonly}[1]{}

\newcommand{\wfourp}{0.885}\newcommand{\lfourp}{0.1475}   
\newcommand{\wtwop}{0.9}\newcommand{\ltwop}{0.1125}       
\newcommand{\wonep}{0.87}\newcommand{\lonep}{0.10875}     
\newcommand{\wabl}{0.9}                                   
\newcommand{\wovl}{0.92}                                  
\newcommand{\wdata}{0.94}                                 
\newcommand{\widobj}{0.92}\newcommand{\lidobj}{0.14956}    
\newcommand{\gidobjs}{0.00219}\newcommand{\gidobjb}{0.00584}

\newcommand{\projectpageline}{\large Project page: \url{https://snap-research.github.io/canvascomposer}}

\definecolor{mediumtealblue}{rgb}{0.0, 0.33, 0.71}
\definecolor{darkpastelgreen}{rgb}{0.01, 0.75, 0.24}
\definecolor{azure}{rgb}{0.0, 0.5, 1.0}

\definecolor{DiagramGreen}{RGB}{46, 188, 132}
\definecolor{DiagramRed}{RGB}{178, 34, 34}

\newcommand{\eg}{e.g.,\xspace}
\newcommand{\ie}{i.e.,\xspace}

\newcommand{\figlabel}{Fig.\xspace}

\newcommand{\seclabel}{Sec.\xspace}

\newcommand{\tablabel}{Tab.\xspace}

\newcommand{\supp}{\textit{Appendix}\xspace}

\definecolor{Highlight}{HTML}{39b54a}  

\usepackage{pifont}
\newcommand{\cmmnt}[1]{}

\newcommand{\rev}[1]{#1}
\newcommand{\revblock}{\relax} 

\usepackage{array}
\usepackage{tabularx}  

\renewcommand{\wfourp}{1.0}\renewcommand{\lfourp}{0.1667}
\renewcommand{\wtwop}{1.0}\renewcommand{\ltwop}{0.125}
\renewcommand{\wonep}{1.0}\renewcommand{\lonep}{0.125}
\renewcommand{\wabl}{1.0}
\renewcommand{\wovl}{1.0}
\renewcommand{\wdata}{1.0}
\renewcommand{\widobj}{1.0}\renewcommand{\lidobj}{0.16256}
\renewcommand{\gidobjs}{0.00238}\renewcommand{\gidobjb}{0.00635}

\renewcommand{\arxivonly}[1]{#1}

\renewcommand{\projectpageline}{\LARGE\href{https://snap-research.github.io/canvascomposer}{\textcolor{magenta}{\fontencoding{OT1}\fontfamily{cmtt}\fontseries{m}\fontshape{n}\selectfont https://snap-research.github.io/canvascomposer}}}

\usepackage{booktabs} 
\usepackage{enumitem}

\usepackage[ruled]{algorithm2e} 

\SetAlFnt{\small}
\SetAlCapFnt{\small}
\SetAlCapNameFnt{\small}
\SetAlCapHSkip{0pt}

\AtEndPreamble{
    \usepackage[capitalize]{cleveref}
    \crefname{section}{Sec.}{Secs.}
    \Crefname{section}{Section}{Sections}
    \Crefname{table}{Table}{Tables}
    \crefname{table}{Tab.}{Tabs.}
}

\AtBeginDocument{%
  }

\setcopyright{none}
\renewcommand\footnotetextcopyrightpermission[1]{}

\newcommand{\methodname}{CanvasComposer}

\begin{document}

\title{CanvasComposer: Personalized Group Photo Generation via a Multi-Reference Canvas}
\author{Gordon Guocheng Qian}
\authornote{Equal contribution.}
\email{gordonqian2017@gmail.com}
\affiliation{%
  \institution{Snap Inc.}
  \city{Palo Alto}
  \country{USA}
}

\author{Ruihang Zhang}
\authornotemark[1]
\email{rhzhang@cs.toronto.edu}
\affiliation{%
  \institution{Snap Inc.}
  \city{Palo Alto}
  \country{USA}
}
\affiliation{%
  \institution{University of Toronto}
  \city{Toronto}
  \country{Canada}
}

\author{Tsai-Shien Chen}
\email{tchen99@ucmerced.edu}
\affiliation{%
  \institution{Snap Inc.}
  \city{Santa Monica}
  \country{USA}
}
\affiliation{%
  \institution{University of California, Merced}
  \city{Merced}
  \country{USA}
}

\author{Yusuf Dalva}
\email{ydalva@vt.edu}
\affiliation{%
  \institution{Snap Inc.}
  \city{Palo Alto}
  \country{USA}
}
\affiliation{%
  \institution{Virginia Tech}
  \city{Blacksburg}
  \country{USA}
}

\author{Anujraaj Goyal}
\email{agoyal@snapchat.com}
\affiliation{%
  \institution{Snap Inc.}
  \city{New York}
  \country{USA}
}

\author{Willi Menapace}
\email{wmenapace@snapchat.com}
\affiliation{%
  \institution{Snap Inc.}
  \city{Santa Monica}
  \country{USA}
}

\author{Ivan Skorokhodov}
\email{iskorokhodov@snapchat.com}
\affiliation{%
  \institution{Snap Inc.}
  \city{Santa Monica}
  \country{USA}
}

\author{Meng Dong}
\email{mdong@snapchat.com}
\affiliation{%
  \institution{Snap Inc.}
  \city{San Diego}
  \country{USA}
}

\author{Arpit Sahni}
\email{asahni@snapchat.com}
\affiliation{%
  \institution{Snap Inc.}
  \city{Santa Monica}
  \country{USA}
}

\author{Daniil Ostashev}
\email{dostashev@snapchat.com}
\affiliation{%
  \institution{Snap Inc.}
  \city{Santa Monica}
  \country{USA}
}

\author{Ju Hu}
\email{erichuju@gmail.com}
\affiliation{%
  \institution{Snap Inc.}
  \city{Santa Monica}
  \country{USA}
}

\author{Mukesh Singhal}
\email{msinghal@ucmerced.edu}
\affiliation{%
  \institution{University of California, Merced}
  \city{Merced}
  \country{USA}
}

\author{Sergey Tulyakov}
\email{stulyakov@snapchat.com}
\affiliation{%
  \institution{Snap Inc.}
  \city{Santa Monica}
  \country{USA}
}

\author{Kuan-Chieh Jackson Wang}
\email{kcjacksonwang@gmail.com}
\affiliation{%
  \institution{Snap Inc.}
  \city{Palo Alto}
  \country{USA}
}

\renewcommand{\shortauthors}{Qian et al.}

\authorsaddresses{}



\begin{abstract}
Existing personalized image generators still struggle to preserve multiple reference identities in natural and coherent multi-human generations. To address these limitations, we present \methodname{}, an interactive framework for personalized group photo generation. Inspired by professional image-editing software, \methodname{} allows users to place reference subjects on a \rev{shared canvas, where each subject keeps its own RGBA cutout of the input}. \rev{This multi-reference} canvas preserves reference content under overlap while providing an intuitive interface for organizing multiple identities\rev{; the subjects remain separate elements on the input canvas, and the model outputs a single personalized and harmonized image}. To keep this representation efficient, transparent latent pruning retains only \rev{tokens from each subject's non-transparent region}, and \rev{cross-reference training} mitigates copy-paste artifacts by learning to harmonize references sampled from different images. Extensive experiments demonstrate that \methodname{} achieves coherent generation and strong identity preservation compared to state-of-the-art methods in multi-human personalized image generation. Project page: \url{https://snap-research.github.io/canvascomposer}
\end{abstract}

%
%



\begin{teaserfigure}
  \centerline{\projectpageline}
  \includegraphics[width=\textwidth]{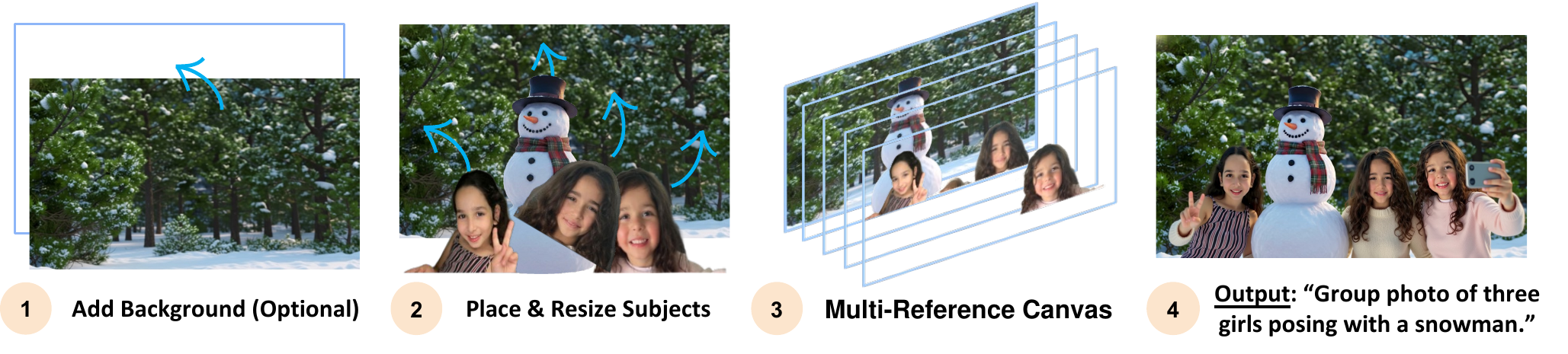}
  \caption{\textbf{\methodname{}} introduces an \textbf{interactive personalization} paradigm for multi-human text-to-image generation. It gives users a Photoshop-like composition experience: reference subjects are placed on a \textbf{\rev{multi-reference} canvas}, with each identity \rev{individually adjustable} through direct canvas manipulation. \methodname{} then harmonizes the \rev{multiple} references into a coherent, high-fidelity group image that follows the prompt.
    }
  \Description{Teaser: a user places cut-out reference subjects on a shared canvas next to an optional background, and CanvasComposer turns that canvas into a single coherent group photo that follows the text prompt.}
  \label{fig:teaser}
\end{teaserfigure}



\maketitle
\pagestyle{plain}
\thispagestyle{plain}

\section{Introduction}\label{sec:intro}

The advent of modern text-to-image (T2I) diffusion models~\citep{rombach2022high} has marked a pivotal moment in digital content creation, enabling the synthesis of complex, high-fidelity images from simple textual descriptions. This breakthrough has spurred a wave of research into personalization, enabling content creation with specified human identities in unseen contexts. Textual Inversion~\citep{gal2022image}, DreamBooth~\citep{ruiz2023dreambooth}, and IP-Adapter~\citep{ye2023ip-adapter} have made significant progress in this direction.

These successful prior works~\cite{ruiz2023dreambooth,ye2023ip-adapter,PuLID} primarily target single-identity personalization. However, personalization is often social: users may want to generate images that include themselves together with friends, family, or other known people, making multi-human personalization a fundamental use case. A successful model must preserve each reference identity, bind each reference to the correct generated person, and render all subjects coherently in a shared scene.

\begin{figure}[t]
\centering
\newcommand{\bglbl}[1]{\makebox[0.3333\linewidth][c]{\scriptsize #1}}%
\bglbl{UniPortrait~\cite{he2024uniportrait}}%
\bglbl{OmniGen2~\cite{wu2025omnigen2}}%
\bglbl{\textbf{\methodname{} (Ours)}}\\[1pt]
\includegraphics[width=1.0\linewidth]{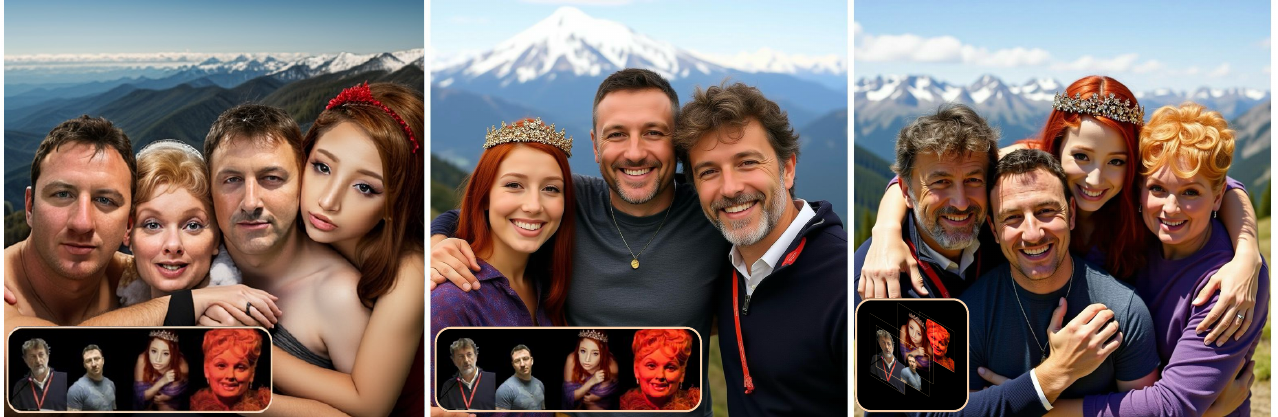}
\caption{\textbf{Motivation for \rev{the multi-reference canvas in} multi-human personalization.}
State-of-the-art multi-human personalization methods such as UniPortrait~\cite{he2024uniportrait} and OmniGen2~\cite{wu2025omnigen2} often struggle in challenging four-person scenes, where reference identities may be swapped, duplicated, or omitted, and conditioning cost grows with the number of subjects. In contrast, \methodname{} introduces a \rev{multi-reference} canvas where users place \rev{and adjust} each reference subject\rev{'s relative location} through direct canvas manipulation, leading to stronger identity preservation, generation quality, and user experience.}
\Description{Three columns of generated four-person group photos. The UniPortrait and OmniGen2 columns show swapped, duplicated, or missing reference identities, while the CanvasComposer column preserves every reference identity in a coherent group photo.}
\label{fig:results:background}
\end{figure}

Existing multi-human personalization methods~\cite{he2024uniportrait,ID-Patch,qian2025composeme,UNO,wu2025omnigen2} typically encode each reference subject into visual conditioning tokens using CLIP-~\cite{CLIP}, face-~\cite{ArcFace,Omni-ID}, or VAE-based encoders~\cite{kingma2013auto}, concatenate the tokens from all subjects, and inject them into a pretrained diffusion model through decoupled cross-attention~\citep{ye2023ip-adapter} or in-context conditioning~\citep{ID-Patch,wu2025omnigen2}. This token-concatenation paradigm faces two key challenges. First, subject association becomes ambiguous as the number of people increases: the model must infer which reference corresponds to which generated person, often causing identity swaps, identity mixing, duplicated subjects, or omissions. Second, the conditioning sequence grows linearly with the number of personalized subjects, increasing computation and memory. As a result, state-of-the-art multi-person personalization methods struggle in four-person settings, as demonstrated in \cref{fig:results:background}. These limitations motivate a new paradigm for \textit{interactive personalization with efficient multi-identity handling}.

To meet this demand, we introduce \textbf{interactive personalization}\rev{, where users directly compose an input canvas for reference-based group photo generation}. Inspired by professional editing tools (e.g., Photoshop), our framework empowers users to act as design directors: they can intuitively compose a scene by placing multiple reference subjects on a \rev{multi-reference} canvas \rev{and adjusting their positions and sizes}, as shown in \figlabel\ref{fig:teaser} (1--2). The canvas \rev{keeps the subjects separated}, allowing users to adjust their relative locations through simple layout manipulation while combining an optional background and multiple identity references in a single \rev{multi-reference} input. The model renders these references into a coherent, high-fidelity, personalized group photo. \rev{Throughout this paper, each subject or background is placed on the canvas as its own reference image with no fixed ordering among them: these references may overlap freely and serve only to specify the personalized subjects for multi-human generation, and the model outputs a single flattened image.}

To render such a template, we propose \textbf{\textit{\methodname{}}}, a generative framework \rev{built on the pretrained FLUX Kontext~\citep{fluxkontext} diffusion transformer and} designed specifically for interactive multi-human T2I generation. The core of \methodname{} is the \textbf{\rev{multi-reference canvas}}, an input representation where each subject \rev{is represented by its own} RGBA \rev{cutout} shown in \figlabel\ref{fig:teaser} (3). \rev{Rather than compositing references into a single flattened input image, the multi-reference canvas keeps subjects separated in the input while preserving the user-edited canvas layout.}
The \rev{multi-reference} representation offers two critical advantages. First, it naturally handles \rev{input-side} occlusion by \rev{keeping each subject as a separate RGBA cutout}, preserving subject content even when \rev{subjects} overlap \rev{in the input canvas}. In contrast, a simple collage-based approach inevitably introduces occlusions when multiple subjects overlap within the input canvas. Such occlusions result in partial identity information loss, and when adjustments are required, such as aligning subjects to match the text prompt or enabling natural interactions, the model inevitably hallucinates the missing regions, leading to reduced identity fidelity (\cref{fig:ablation}). Second, we introduce a \textbf{transparent latent pruning} strategy demonstrated in \cref{fig:pipeline}, which extracts and concatenates only valid (non-transparent) tokens from \rev{each subject}. By discarding empty transparent regions, this design makes the conditioning length depend on non-transparent reference content rather than the number of \rev{subjects placed on the canvas}, eliminating the linear per-subject token growth of prior concatenation-based methods and improving efficiency for multi-human generation.

\begin{figure*}[t]
    \centering
    \includegraphics[width=\wdata\linewidth, trim={0cm 0.5cm 0cm 0cm}, clip]{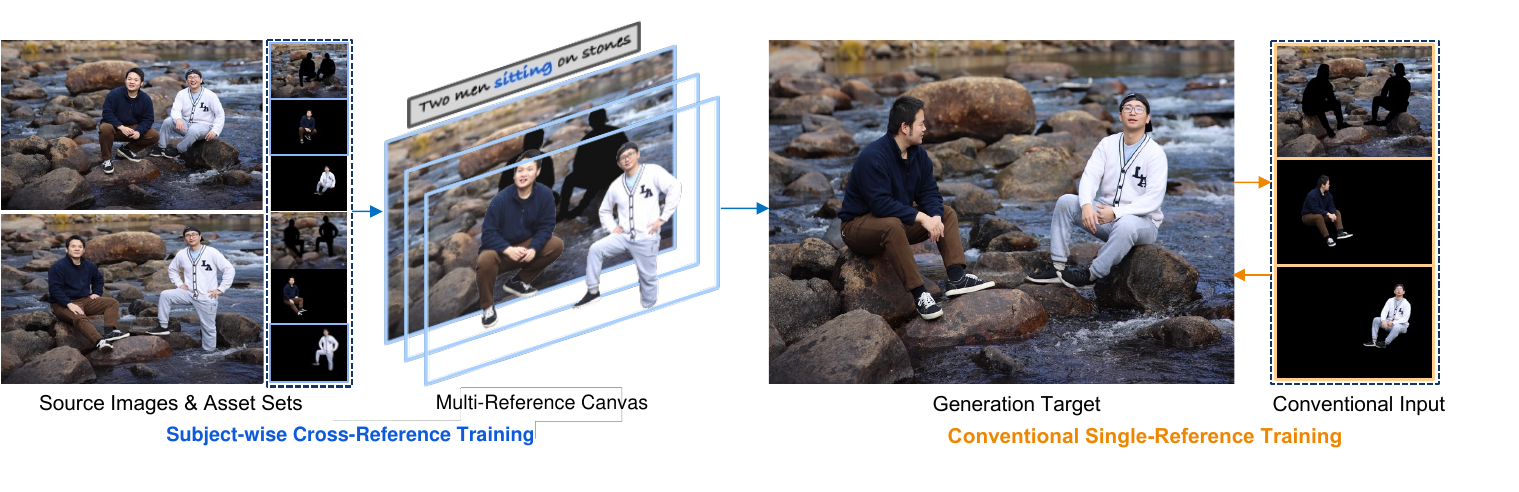}
    \caption{\textbf{\rev{Subject-wise Cross-Reference} Sampling Strategy.}
    During training, \methodname{} constructs the \rev{multi-reference} canvas by sampling \rev{each subject} from a random source image within the same identity group, while the target image is a different sample from that group. This strategy intentionally introduces mismatches both among \rev{the input images} and between the inputs and the target, encouraging \methodname{} to learn adaptive alignment to the prompt and context during inference. In contrast, conventional single-reference training can overfit to pixel-level correspondences, often leading to outpainting or copy-paste artifacts (see examples in \cref{fig:ablation}).}
    \Description{Training data diagram: for each subject the multi-reference canvas is built from a random source image of that identity while the target is a different image of the same identity group, so inputs and target never match pixel for pixel.}
    \label{fig:data}
\end{figure*}

To train \methodname{}, the conventional training pipeline constructs such a canvas by cropping human segments directly from the target image \rev{to place on the canvas} (\cref{fig:data} right). However, this approach often results in undesirable copy-paste artifacts due to pixel-level correspondence between the input and target.
To address this, we introduce \textbf{\rev{subject-wise cross-reference training}} (\cref{fig:data} left). We curate a multi-image-per-scene dataset in which each identity group contains multiple images. During training, an image is randomly chosen as the target, while the corresponding humans and background in the \rev{multi-reference} canvas are sampled from other source images within the same image set. Each sampled human is fitted to the target bounding boxes, with additional geometric and lighting augmentations to introduce intentional mismatches between inputs and targets. The \rev{subject-wise cross-reference training} strategy encourages the model to disentangle redundant factors such as pose and illumination \rev{across subjects}, thereby harmonizing multiple inputs from diverse contexts into a single coherent image during inference.

Our main \textbf{contributions} are as follows:
\begin{itemize}[leftmargin=*,noitemsep,topsep=0pt,parsep=0pt,partopsep=0pt]
\item We propose an \textbf{interactive personalization paradigm} that allows users to place multi-human references in full-body, portrait, or cropped-head forms on a \rev{multi-reference} canvas.

\item We introduce the \textbf{\rev{multi-reference canvas}}, a simple and effective \rev{input} representation that handles occlusions in multi-human personalization and improves efficiency through our \textbf{transparent latent pruning} strategy.

\item We present \textbf{\rev{the subject-wise cross-reference training strategy}}, which effectively disentangles redundant visual information (e.g., poses) and mitigates copy-paste artifacts.

\item We develop \textbf{\methodname{}}, a practical framework for interactive multi-human personalized generation. Extensive experiments demonstrate that \methodname{} achieves state-of-the-art identity preservation and generation quality across multi-human personalization benchmarks.
\end{itemize}

\section{Related Work}\label{sec:related}

\paragraph{Personalized Generation}
Personalization methods have evolved from costly per-concept tuning~\citep{gal2022image,nitzan2022mystyle,ruiz2023dreambooth,custom-diffusion} to recent adapter-based solutions~\citep{ye2023ip-adapter,han2024emma,li2023photomaker,wang2024instantid,gal2024lcm,PuLID,Omni-ID,patashnik2025nested,goyal2025preventing}, which enable efficient personalization while keeping the base diffusion model frozen.
For multi-subject personalization, optimization-based approaches require additional concept disentanglement~\citep{custom-diffusion,avrahami2023break,tokenverse} or training multiple LoRAs~\citep{po2023orthogonal,kong2024omg,dalva2025lorashop}.
Optimization-free approaches~{\NoHyper\citep{xiao2024fastcomposer,wang2024moa,instantfamily,StoryMaker,videoalchemist}\endNoHyper}, including recent works such as UniPortrait~\cite{he2024uniportrait}, ID-Patch~\citep{ID-Patch}, ComposeMe~\citep{qian2025composeme}, and MS-Diffusion~\citep{MS-Diffusion}, rely on lightweight adapters but become increasingly expensive as the number of subject tokens grows.
Recent in-context models~\citep{mi2025thinkdiff,OmniGen,UNO,DreamO, comanici2025gemini,qwen-image} support multiple concept combinations, but they provide limited interactivity and their cost still grows with the number of subjects.
\arxivonly{Ar2Can~\citep{ar2can} likewise targets multi-human generation with a canvas, but its canvas is a layout predicted by an \emph{Architect} module and then rendered by an \emph{Artist} module trained with GRPO under a spatially grounded face-matching reward, whereas our canvas is composed by the user from RGBA reference images and preserves each reference under overlap.}
Our \methodname{} instead advances this line of work by introducing a \rev{multi-reference} canvas that supports interactive subject placement, occlusion-aware reference preservation, and efficient multi-human personalization.

\paragraph{Structural and \rev{Compositional} Conditioning}
A broad range of conditioning mechanisms has been proposed to guide T2I models with additional visual or structural inputs.
ControlNet~\citep{zhang2023adding} and T2I-Adapter~\citep{mou2023t2i} are pioneering works that inject structural cues through auxiliary components, allowing the diffusion model to incorporate signals such as pose or depth.
GLIGEN~\citep{li2023gligen}, LayoutDiffusion~\citep{zheng2023layoutdiffusion}, and their follow-ups~\citep{xiong2024groundingbooth, song2023objectstitch,chen2023anydoor,jiang2024hico,zhang2025creatidesign} fine-tune diffusion models to interpret bounding boxes or segmentation masks for structured generation.
Training-free approaches such as NoiseCollage~\citep{noisecollage}, GrounDiT~\citep{lee2024groundit}, and Bounded Attention~\citep{dahary2024yourself} manipulate noise or attention maps at inference. In terms of layered image generation, CollageDiffusion~\citep{sarukkai2023collage} and LayerDiffusion~\cite{li2023layerdiffusion} are leading works, which require per-layer optimization through textual inversion~\cite{gal2022image} to keep layers separated. SceneDiffusion~\citep{ren2024move} extends LayerDiffusion to remove the optimization stage but still requires per-layer generation for image editing. \rev{Latent transparency~\cite{zhang2024transparent} and RGBA-instance generation~\citep{fontanella2024rgba} instead produce transparent layers on the \emph{output} side, and LaRender~\citep{larender2025} controls output occlusion training-free by volume-rendering prompt-described objects in latent space; none of these condition on reference images for identity personalization. In contrast to these box-, attention-, or RGBA-\emph{output}-based mechanisms, \methodname{} takes \rev{a separate RGBA image per subject} as \rev{multi-reference} \emph{input} conditioning directly, preserving reference identities even when \rev{these input images} overlap; concurrent layered-generation efforts are complementary to this input-side design.}
LazyDiffusion~\citep{nitzan2024lazy} is also related in spirit because it reduces diffusion-transformer computation for localized editing. However, it prunes target denoising tokens inside a user-specified mask, whereas \methodname{} prunes input conditioning tokens from transparent regions across \rev{multiple reference images}.
The most relevant work to ours is ID-Patch~\citep{ID-Patch}, a feedforward personalized T2I model trained on collaged face inputs, where each patch consists of projected face features optionally overlaid with pose. However, ID-Patch is restricted to face-only generation and suffers from occlusion issues as well as copy-paste artifacts. 
In contrast, \methodname{} organizes full-body, portrait, or cropped-head references \rev{as separate input images}, preserves overlapping inputs, and effectively mitigates copy-paste artifacts via the proposed \rev{subject-wise cross-reference} training.

\section{\methodname{}}\label{sec:method}

\subsection{\rev{Multi-Reference Canvas}}
\methodname{} is a multi-human text-to-image generation framework that offers an interactive personalization experience, enabling users to specify the appearance of multiple subjects (humans and an optional background) through \rev{separate} visual references. Concretely, the framework conditions a pretrained diffusion model on two inputs: (1) a text prompt that specifies global image content and high-level semantics and (2) a \textbf{\rev{multi-reference canvas}} that \rev{keeps each subject as a separate reference element}. This design supports intuitive canvas layout manipulation while keeping identity-specific guidance separated in a globally coherent image.

Illustrated in the second column of \cref{fig:data}, the \textbf{\rev{multi-reference canvas}} is represented by a set of RGBA \rev{images} $L = \{l_1, \cdots, l_N\}$, \rev{one per subject or background element}, where $N$ \rev{is the number of such elements}. Each RGBA \rev{image} $l_i$ represents one element, either a human segment or a background image. The RGB channels provide visual reference for the element, while the alpha channel defines valid regions. This mask is used to identify valid tokens, while invalid tokens are pruned out for efficiency, as detailed in \seclabel\ref{sec:method_pipeline}. When an optional background is provided, regions covered by foreground subjects are removed from the background during both training and inference, letting the diffusion model harmonize the final image according to the text prompt.

\rev{Each subject or background is placed separately on the canvas with no fixed ordering among them; a per-element index only keeps overlapping references separable (\seclabel\ref{sec:method_pipeline}) and carries no depth semantics. The alpha channel is used to prune input tokens for efficiency, and the output is a single flattened image, the final personalized group photo.}

\subsection{\methodname{} Pipeline}
\label{sec:method_pipeline}

\begin{figure}
    \centering
    \includegraphics[width=1.0\linewidth]{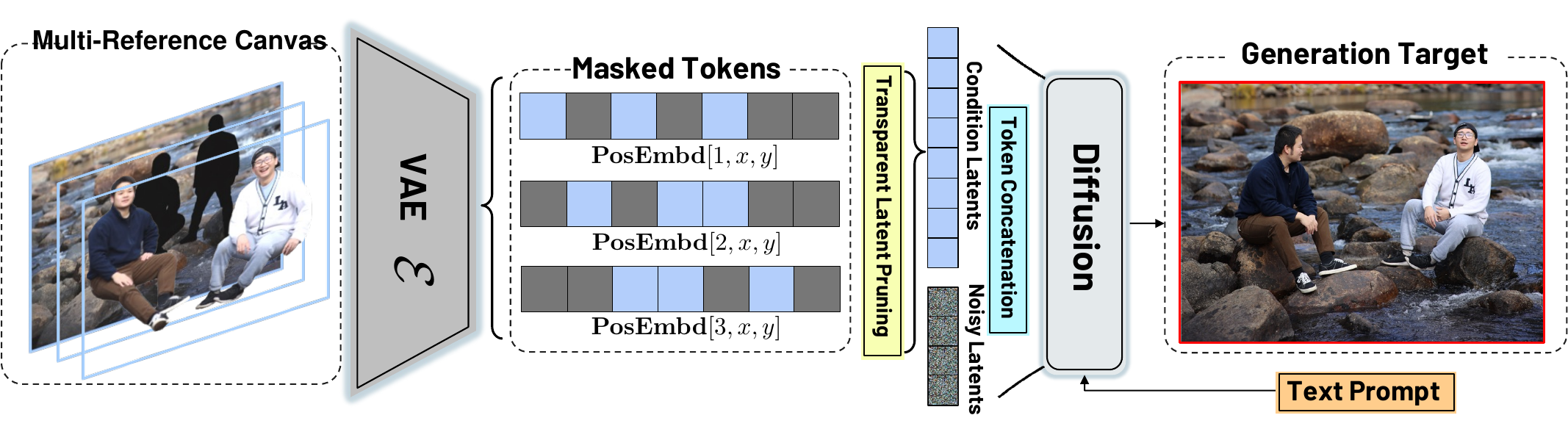}
\caption{\noindent\textbf{\methodname{} Pipeline.} \methodname{} conditions a diffusion model on a text prompt and a {\rev{multi-reference canvas}}. Each RGBA \rev{image} is encoded into latent tokens with a VAE and assigned a unique \rev{index}~$j$, yielding 3D positional embeddings $\big[j, x, y\big]$ that \rev{tell overlapping subjects apart}. {Transparent latent pruning} keeps only non-transparent conditioning tokens and discards transparent regions (\textcolor[HTML]{666666}{gray boxes}), reducing the conditioning sequence for efficient personalized generation.}
    \Description{Pipeline diagram: a text prompt and a multi-reference canvas are encoded; each RGBA reference image becomes latent tokens with a unique index and three-dimensional positional embeddings, transparent latent pruning discards the transparent tokens, and the remaining tokens condition the diffusion model that generates the group photo.}
    \label{fig:pipeline}
\end{figure}

\methodname{} builds on a pretrained latent-based diffusion transformer (DiT)~\citep{Peebles2023DiT}, as illustrated in \figlabel\ref{fig:pipeline}. 
To preserve the pretrained model's capacity as much as possible, \methodname{} treats the \rev{multi-reference} canvas as pruned in-context visual tokens (reference tokens processed jointly with the noisy image tokens) and reuses the DiT's native self-attention and 3D Rotary Position Embedding (RoPE)~\citep{su2024roformer} interface \rev{for element-aware conditioning}, without adding a decoupled attention branch or per-subject adapter.

\paragraph{\rev{Per-Element Latent Extraction}} For each input \rev{element} $l_i \in L$, we first encode the RGB content using the pretrained VAE encoder to obtain \rev{its} latents $z_i = \mathcal{E}(l_i^{\text{RGB}}) \in \mathbb{R}^{H' \times W' \times D}$ where $\mathcal{E}$ is the VAE encoder, $H'$ and $W'$ are the height and width of the latent grid, and $D$ is the feature dimension.

\paragraph{\rev{Positional Embeddings for Canvas Elements}}
We adopt a simple yet effective positional embedding scheme to encode both the 2D latent-grid coordinate and \rev{which element each latent token belongs to}, following the 3D RoPE parametrization of FLUX Kontext~\citep{fluxkontext}. For every \rev{element's} latent $z_i$, we construct a 3D positional embedding
\begin{equation}
    \text{pos}_i = [j_i, x, y] \in \mathbb{R}^3,
\end{equation}
where $(x,y)$ are the coordinates in the latent grid and $j_i \in \{0,1,\ldots,N\}$ \rev{indexes which canvas element a token belongs to}. FLUX Kontext already uses this 3D RoPE interface; we reuse its parametrization unchanged \rev{(full formulation in \supp)}. The original model sets $j_i=0$ for image tokens; we reserve this index for the noisy latent tokens of the base diffusion model and assign $j_i \ge 1$ to \rev{each user-placed element on the canvas}. This design preserves the pretrained positional interface of the base model while allowing attention layers to \rev{tell overlapping canvas elements apart}. \rev{We emphasize that the index $j_i$ acts purely as an \emph{input}-side identity-separation signal: it follows the segmenter's confidence-ranked mask order and carries no depth or z-ordering semantics. Accordingly, \methodname{} resolves occlusion within the \emph{input} canvas by preserving each reference's pixels when subjects overlap (\cref{tab:overlap_ablation}); explicit control over occlusion or depth ordering in the \emph{output} image is left as future work, \eg by integrating training-free occlusion control~\citep{larender2025} or by training with depth-conditioned inputs, which would require depth estimation during data generation.} Such positional embeddings are injected into the attention blocks through RoPE~\cite{su2024roformer}, following the design of the base model.

\paragraph{Transparent Latent Pruning} To improve the efficiency of multi-subject personalization, we introduce the transparent latent pruning strategy that selectively retains the latent tokens from valid non-transparent regions according to the alpha channel, while discarding the rest. Concretely, for \rev{each element's} alpha channel $l_i^{\alpha}$, we first downsample it to the latent resolution:
\begin{equation}
    \alpha_i^{\text{latent}} = \text{NearestResize}(l_i^{\alpha}) \in \mathbb{R}^{H' \times W'}.
\end{equation}
We then apply alpha-based masking to the latent tokens $z_i$, keeping only those in regions with non-zero alpha values:
\begin{equation}
    z_i^{\text{valid}} = \text{Concat}(\{z_i(x, y) | \alpha_i^{\text{latent}}(x, y) > 0.5\}),
\end{equation}
where $z_i(x, y)$ and $\alpha_i^{\text{latent}}(x, y)$ denote the latent tokens and their alpha values at latent-grid coordinates $(x, y)$.

Training and inference use the identical token-selection rule. Transparent RGB regions are not used as explicit conditioning tokens: after VAE encoding, tokens whose downsampled alpha is transparent are discarded before the DiT receives the conditioning sequence. Although VAE tokens near boundaries may contain local context from neighboring RGB pixels, the model is fine-tuned with this exact masking formulation, and only retained tokens are exposed to the DiT. Practically, we see no boundary artifacts or identity leakage from transparent regions, so explicit partial or masked convolutions are unnecessary for our setting.

\rev{Unlike existing personalization methods~\citep{qian2025composeme,ID-Patch} that allocate a fixed-length token block per subject on a full-canvas crop and thus scale linearly with the subject count, \textit{the transparent latent pruning strategy makes the length of the token sequence proportional to the non-transparent content placed on the shared canvas, rather than to the number of subjects placed on the canvas}. Because users place and resize segmented subjects on a single fixed canvas, per-subject footprints shrink as more subjects are added, thus yielding substantial efficiency improvements.}

\paragraph{\rev{Combining the Conditioning Tokens}} Finally, we construct the conditional latents by aggregating the pruned latents from \rev{every element on the canvas}:
$z_{\text{cond}} = \text{Concat}(z_1^{\text{valid}}, z_2^{\text{valid}}, \ldots, z_N^{\text{valid}})$,
which is further concatenated with the noisy image latents $z_t$ to form the latent input of the DiT model, as shown in \figlabel\ref{fig:pipeline}. 

\begin{figure*}[h!]
\centering
\newcommand{\fourlbl}[1]{\makebox[\lfourp\linewidth][c]{\scriptsize\textbf{#1}}}%
\fourlbl{UniPortrait}\fourlbl{ID-Patch}\fourlbl{UNO}%
\fourlbl{OmniGen2}\fourlbl{Ours}\fourlbl{Inputs}\\[1pt]
\includegraphics[width=\wfourp\linewidth]{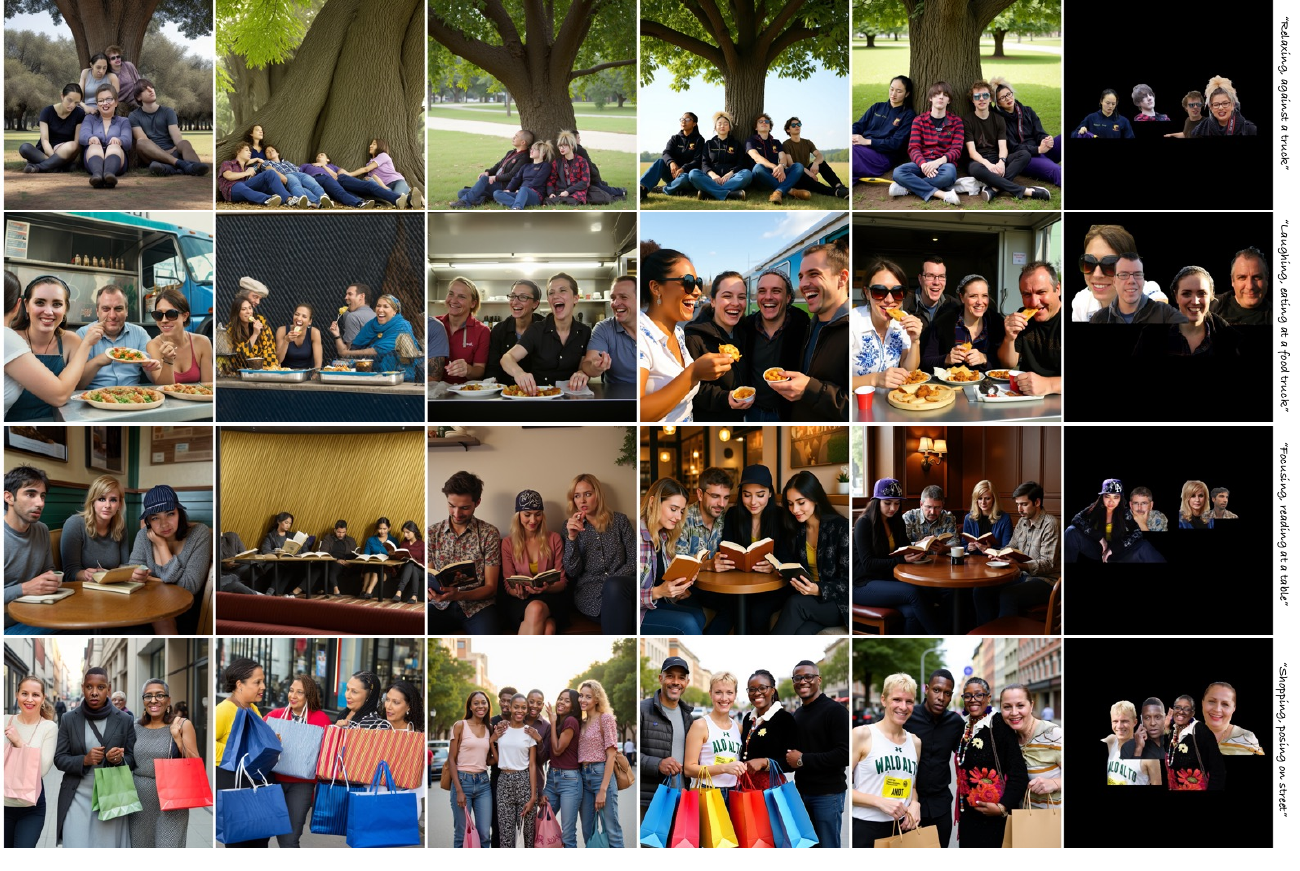}
\caption{\textbf{Qualitative Comparison in Four-person (4P) Personalization.} State-of-the-art baselines often fail to preserve all reference identities, causing identity swaps, distortions, omissions, or duplicated subjects. {\methodname{}} consistently generates high-fidelity results that preserve all identities in challenging multi-person scenes. The ``Inputs'' column shows the \rev{multi-reference-canvas} visualization for \methodname{}, whereas all baselines receive individual images as required by their input formats throughout all benchmarks.}
\Description{A grid of four-person group photos from four baselines and from CanvasComposer, with the multi-reference canvas shown in the last column. The baseline rows show swapped, distorted, missing, or duplicated people, while the CanvasComposer rows keep all four reference identities.}
\label{fig:results:4p}
\end{figure*}

\subsection{\methodname{} Training}\label{sec:method:training}

\paragraph{\rev{Subject-wise Cross-Reference Data Sampling Strategy}}
\methodname{} training employs a \emph{\rev{subject-wise cross-reference}} data sampling strategy, as illustrated in \figlabel\ref{fig:data}, to mitigate copy-paste artifacts. Our training requires a multi-image-per-scene dataset (\seclabel\ref{sec:exp_setup}), where each scene contains several images depicting the same human subjects under different poses, viewpoints, and lighting conditions.
For each training sample, we randomly select an image as the ground-truth target, denoted $I^{\text{target}}$, and use the remaining images in the same scene as a pool of cross-reference source images.
For every subject $i$ in the target, we collect an asset set $\mathcal{A}_i$ consisting of its segments across the remaining source images within the scene.

To construct the input \rev{multi-reference} canvas $L$, \rev{we build it one subject at a time}. For each subject $i$, we randomly sample one asset $a_i \in \mathcal{A}_i$ and fit it to the subject's bounding box in $I^{\text{target}}$, \rev{giving a distinct RGBA image} $l_i$. This training canvas uses target boxes only for supervised data construction, while sourcing appearance cues for each subject from different images within the same scene.

This \rev{subject-wise cross-reference} construction produces training inputs where different subjects originate from different source images, rather than from a single reference frame as in the conventional training paradigm (see \cref{fig:data} right). Our strategy encourages \methodname{} to mitigate copy-paste artifacts by integrating cross-image identity cues and to generalize across diverse poses, expressions, viewpoints, and contexts.

\paragraph{\rev{Canvas-Conditioned Finetuning}}
We adapt the pretrained DiT with lightweight LoRA finetuning using a flow matching loss~\citep{lipman2022flow} with \rev{multi-reference} canvas conditioning:
\begin{equation}
    \mathcal{L}_{\text{cond}} 
    = \mathbb{E}_{\substack{t\sim\,\mathcal{U}(0, 1), z_0,\\z_1, z_{\text{cond}}, P}}
    \left[ \, \big\| v_{\theta}(z_{t}, t, z_{\text{cond}}, P) - (z_1 - z_0) \big\|^2 \, \right]
\end{equation}
where $v_{\theta}(\cdot)$ is the predicted velocity, $z_1$ and $z_0$ are the latents of the target image and the sampled noise, $z_t$ is the noisy latents at timestep $t$ of the target image $I^{\text{target}}$, $z_{\text{cond}}$ is the conditional latents of our \rev{multi-reference} canvas $L$, and $P$ is the text prompt.

\section{Experiments}\label{sec:exp}

\begin{figure*}[t]
\centering
\newcommand{\twolbl}[1]{\makebox[\ltwop\linewidth][c]{\scriptsize\textbf{#1}}}%
\twolbl{UniPortrait}\twolbl{StoryMaker}\twolbl{ID-Patch}\twolbl{UNO}%
\twolbl{DreamO}\twolbl{OmniGen2}\twolbl{Ours}\twolbl{Inputs}\\[1pt]
\includegraphics[width=\wtwop\linewidth]{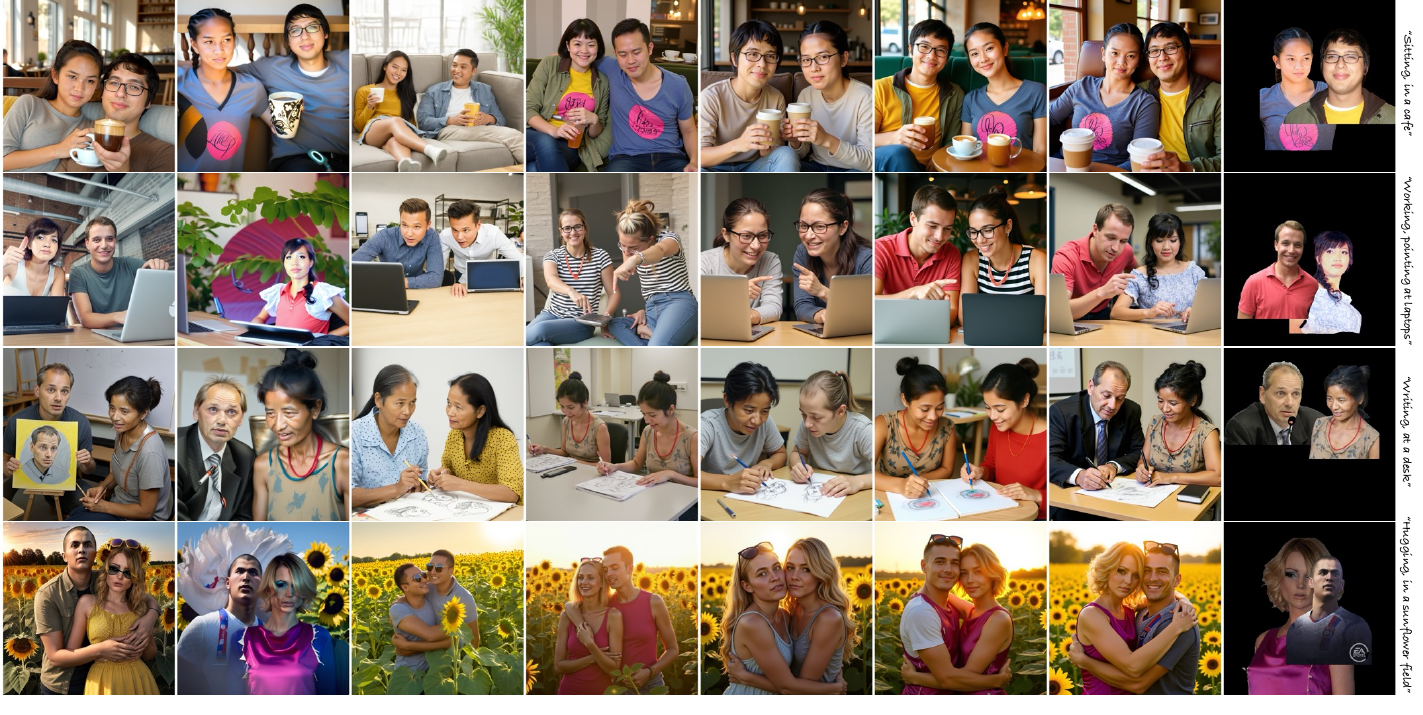}
\caption{\textbf{Qualitative Comparison in Two-person (2P) Personalization.}
Competing methods can miss one identity, mix identities between subjects, or copy the references too rigidly.
\methodname{} produces coherent two-person results that preserve both reference identities.}
\Description{A grid of two-person group photos produced by six baselines and by CanvasComposer, with the reference inputs in the last column. Baselines drop one of the two identities or blend the two faces together, while CanvasComposer preserves both identities in a single coherent photo.}
\label{fig:results:2p}
\end{figure*}

\begin{table}[t]
\centering
\caption{\textbf{Quantitative comparison.}
\methodname{} consistently achieves superior identity preservation (ArcFace~\cite{ArcFace}) in the 4P and 2P setups while maintaining strong aesthetic quality (HPSv3~\cite{ma2025hpsv3}) and prompt alignment (VQAScore~\cite{VQAScore}).
We also conduct a pairwise user study detailed in \supp, where ``Ours Win Rate (\%)'' denotes the fraction of comparisons in which our method is preferred over a baseline. Such user studies are important for personalization evaluation as automatic metrics can be noisy or biased. \rev{Across the 4P and 2P benchmarks, \methodname{} is preferred over every baseline: winning over 84\% of comparisons against personalization methods and over 60\% against recent image-editing models (FLUX Kontext, Overlay Kontext, Qwen-Image-Edit, and Nano-Banana).}
}
\label{tab:numbers}
\resizebox{\linewidth}{!}{%
\begin{tabular}{lcccc}
\toprule
\textbf{Method} & \textbf{ArcFace}~$\uparrow$ & \textbf{HPSv3}~$\uparrow$ & \textbf{VQAScore}~$\uparrow$ & \textbf{Ours Win Rate (\%)}~$\uparrow$ \\
\midrule
\multicolumn{5}{c}{\textbf{4P Personalization}} \\
UniPortrait~\cite{he2024uniportrait} & 0.309 & 12.4 & 0.786 & 95.6 \\
ID-Patch~\cite{ID-Patch} & 0.082 & 7.13 & 0.785 & 98.0 \\
UNO~\cite{UNO} & 0.077 & 12.2 & 0.840 & 94.3 \\
OmniGen2~\cite{wu2025omnigen2} & 0.086 & \textbf{13.0} & 0.805 & 97.8 \\
\rev{FLUX Kontext~\cite{fluxkontext}} & \rev{0.217} & \rev{12.8} & \rev{0.869} & \rev{92.2} \\
\rev{Overlay Kontext~\cite{ilkerzgi2025overlay}} & \rev{0.251} & \rev{11.2} & \rev{0.828} & \rev{93.0} \\
\rev{Qwen-Image-Edit~\cite{qwen-image}} & \rev{0.223} & \rev{\textbf{13.0}} & \rev{\textbf{0.895}} & \rev{68.9} \\
\rev{Nano-Banana~\cite{comanici2025gemini}} & \rev{0.434} & \rev{10.4} & \rev{0.826} & \rev{60.6} \\
\textbf{\methodname{} (Ours)} & \textbf{0.533} & 12.5 & 0.840 & Ref. \\
\midrule
\multicolumn{5}{c}{\textbf{2P Personalization}} \\
UniPortrait~\cite{he2024uniportrait} & 0.460 & \textbf{13.2} & 0.812 & 86.9 \\
StoryMaker~\cite{StoryMaker} & 0.542 & 4.97 & 0.523 & 92.8 \\
ID-Patch~\cite{ID-Patch} & 0.121 & 10.1 & 0.858 & 98.4 \\
UNO~\cite{UNO} & 0.072 & 11.2 & 0.870 & 93.1 \\
DreamO~\cite{DreamO} & 0.443 & 12.4 & \textbf{0.877} & 93.4 \\
OmniGen2~\cite{wu2025omnigen2} & 0.121 & 12.8 & 0.828 & 84.7 \\
\textbf{\methodname{} (Ours)} & \textbf{0.547} & 11.6 & 0.865 & Ref. \\
\bottomrule
\end{tabular}}
\end{table}

\subsection{Experimental Setup}
\label{sec:exp_setup}

\paragraph{Training Dataset Curation}
Our training set comprises 32M in-house images across 6M scenes, with a focus on human subjects\footnote{Our internal dataset cannot be released due to internal data policies. Following~\cite{Omni-ID}, a similar dataset can be curated by sampling from public videos.}. The training images are paired with standard captions, which provide the text prompt used to condition global scene content during training. We filter the data to ensure each scene contains at most 4 identities and to exclude low-resolution, low-quality faces. To construct our \rev{multi-reference} training data, we apply \rev{an in-house, YOLO11-style~\citep{yolo11} human instance segmenter (a public YOLO11-seg model can serve as a substitute)} to extract each human \rev{as a separate RGBA cutout} and retain the remaining pixels as background. When constructing the input \rev{multi-reference} canvas in each training step, we apply \rev{per-subject} augmentations including random geometric jitter, color jitter with brightness $\pm0.5$, contrast in $[0.5,1.0]$, saturation $\pm0.5$, no hue shift, and object-level Gaussian blur and grayscale augmentation with probability $0.1$ each.

\paragraph{Training Details}
We train a LoRA with a rank of 512 on all attention layers of the frozen FLUX Kontext~\citep{fluxkontext} using the AdamW optimizer~\citep{AdamW}. The original attention projection matrices are $3072{\times}3072$, making rank 512 a compact but expressive adaptation. The model is trained for 200K iterations with a constant learning rate of $1{\times}10^{-4}$, a total batch size of 32, and a $512{\times}512$ resolution. The entire training took 4 days on 4 nodes, each with 8 A100 GPUs.

\paragraph{Evaluation Details}
We evaluate at $1024{\times}1024$ resolution using 128 images from FFHQ-in-the-wild~\citep{StyleGAN} as identity inputs. FFHQ is a public, single-frame dataset and is not included in training. There are $32$ prompts for each benchmark. All evaluations are conducted with $28$ denoising steps for \methodname{}, without any post-processing. See \supp for evaluation details.

\subsection{Baseline Comparisons}

\paragraph{Four-Person (4P) Personalization}
Most existing human-centric personalization methods~\cite{ye2023ip-adapter, StoryMaker, DreamO, qian2025composeme} struggle in four-person settings because their computation and memory grow with the number of subjects. In contrast, \methodname{}, enabled by our \rev{multi-reference} canvas and transparent latent pruning, supports multiple subjects with substantially lower conditioning cost.

We compare \methodname{} in the 4P setting against recent multi-subject personalization methods: UniPortrait~\citep{he2024uniportrait}, ID-Patch~\citep{ID-Patch}, UNO~\citep{UNO}, and OmniGen2~\citep{wu2025omnigen2}. \methodname{} demonstrates stronger performance in this task.
As shown in \figlabel~\ref{fig:results:4p}, \methodname{} generates high-quality images that preserve all reference identities in challenging multi-person scenes.

Quantitatively, as reported in \tablabel~\ref{tab:numbers}, \methodname{} achieves \rev{the highest identity preservation by a large margin while maintaining competitive text-image alignment and aesthetic quality}, as assessed by ArcFace~\cite{ArcFace}, VQAScore~\cite{VQAScore}, and HPSv3~\cite{ma2025hpsv3}, respectively. \rev{Our user study reveals a strong preference towards \methodname{}, where participants favored it over all state-of-the-art 4P personalization baselines in over 96\% of pairwise comparisons on average.}
\rev{We further compare against recent image-editing approaches, FLUX Kontext~\cite{fluxkontext}, Overlay Kontext~\cite{ilkerzgi2025overlay}, Qwen-Image-Edit~\cite{qwen-image}, and Nano-Banana~\cite{comanici2025gemini}, on this challenging 4P task, as these methods have recently emerged as strong baselines for multi-subject personalization. As shown in \tablabel~\ref{tab:numbers}, \methodname{} outperforms all of them, achieving the highest identity preservation and the strongest user preference (see \supp for qualitative comparisons and details).}
\rev{We note an input asymmetry in these comparisons: \methodname{} and the editing baselines receive the composited canvas that encodes the intended layout, whereas prior personalization baselines accept only individual reference images without spatial layout control.}

\begin{figure*}[t]
\centering
\newcommand{\onelbl}[1]{\makebox[\lonep\linewidth][c]{\scriptsize\textbf{#1}}}%
\onelbl{UniPortrait}\onelbl{StoryMaker}\onelbl{ID-Patch}\onelbl{UNO}%
\onelbl{DreamO}\onelbl{OmniGen2}\onelbl{Ours}\onelbl{Inputs}\\[1pt]
\includegraphics[width=\wonep\linewidth,trim=0 283.5 0 0,clip]{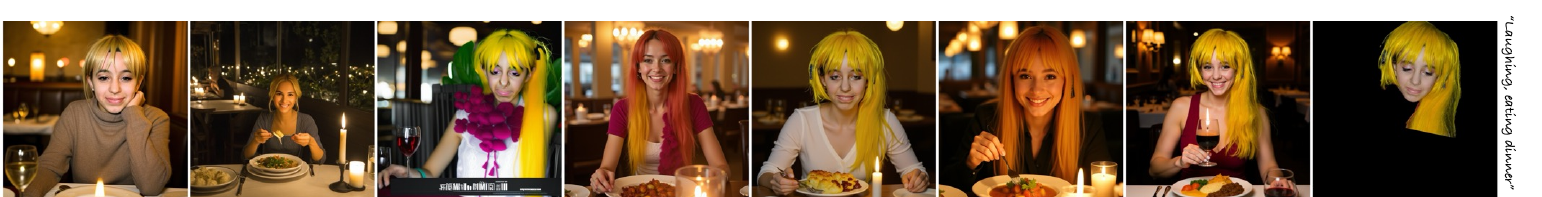}\\
\includegraphics[width=\wonep\linewidth,trim=0 0 0 189,clip]{figures/src/1p.pdf}
\caption{\textbf{Qualitative Comparison in Single-Person (1P) Personalization.}
State-of-the-art methods often fail to decouple identity from pose and expression, leading to copy-paste artifacts, whereas \methodname{} remains faithful to the reference identity while following diverse prompts (\eg \emph{smiling} and \emph{laughing}).}
\Description{A grid of single-person portraits generated by seven baselines and by CanvasComposer, with the reference input shown in the last column. Baseline portraits copy the pose and expression of the reference, whereas the CanvasComposer portraits keep the identity while following prompts such as smiling and laughing.}
\label{fig:results:1p}
\end{figure*}

\paragraph{Two-Person (2P) Personalization}
Most existing personalization methods are designed specifically for up to 2P personalization. Here, we evaluate \methodname{} for the 2P personalization task against the aforementioned 4P personalization methods and recent 2P personalization approaches including StoryMaker~\citep{StoryMaker} and DreamO~\citep{DreamO}.
As illustrated in \figlabel\ref{fig:results:2p}, previous approaches again frequently struggle in two-person generation. They may omit one subject, duplicate identities, or fail to retain facial identity information, resulting in unnatural interactions or low-quality outputs. In contrast, our method consistently produces high-fidelity outputs where both identities are faithfully preserved.
These qualitative trends are corroborated by quantitative evaluations in \tablabel\ref{tab:numbers}. \methodname{} is strongly preferred by users and achieves the best identity preservation among all baselines.

\begin{figure}
    \centering
    \includegraphics[width=1.0\linewidth]{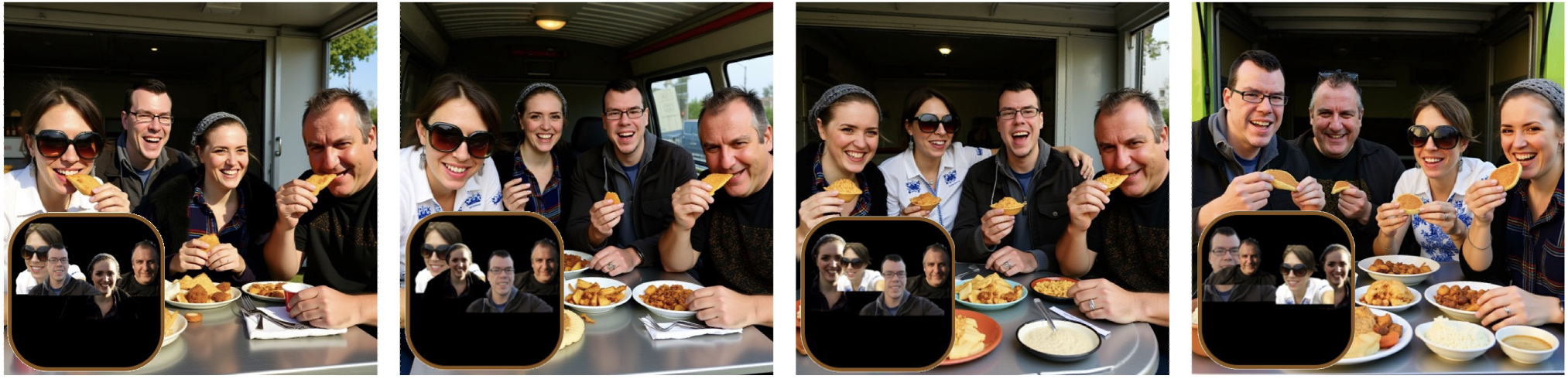}

\caption{\textbf{Effect of \rev{Multi-Reference} Canvas Layout.} The \rev{multi-reference} canvas defines the relative spatial arrangement of subjects while the text prompt controls pose, interaction, and scene context.}

    \Description{Three generated group photos from the same reference subjects arranged differently on the canvas, showing that the canvas layout sets the spatial arrangement while the prompt controls pose, interaction, and scene.}
    \label{fig:app-reorder}
\end{figure}

\paragraph{Single-Person (1P) Personalization}
Although our primary focus is multi-subject personalization, \methodname{} is equally effective for single-subject scenarios. \rev{As shown in \cref{fig:results:1p}, competing approaches tend to inject the reference identity with limited pose or expression variability, whereas \methodname{} faithfully follows diverse prompts. See \supp for the quantitative comparisons and the comparisons with single-subject personalization methods.}

\begin{figure*}[t]
    \centering
\includegraphics[width=\wabl\linewidth]{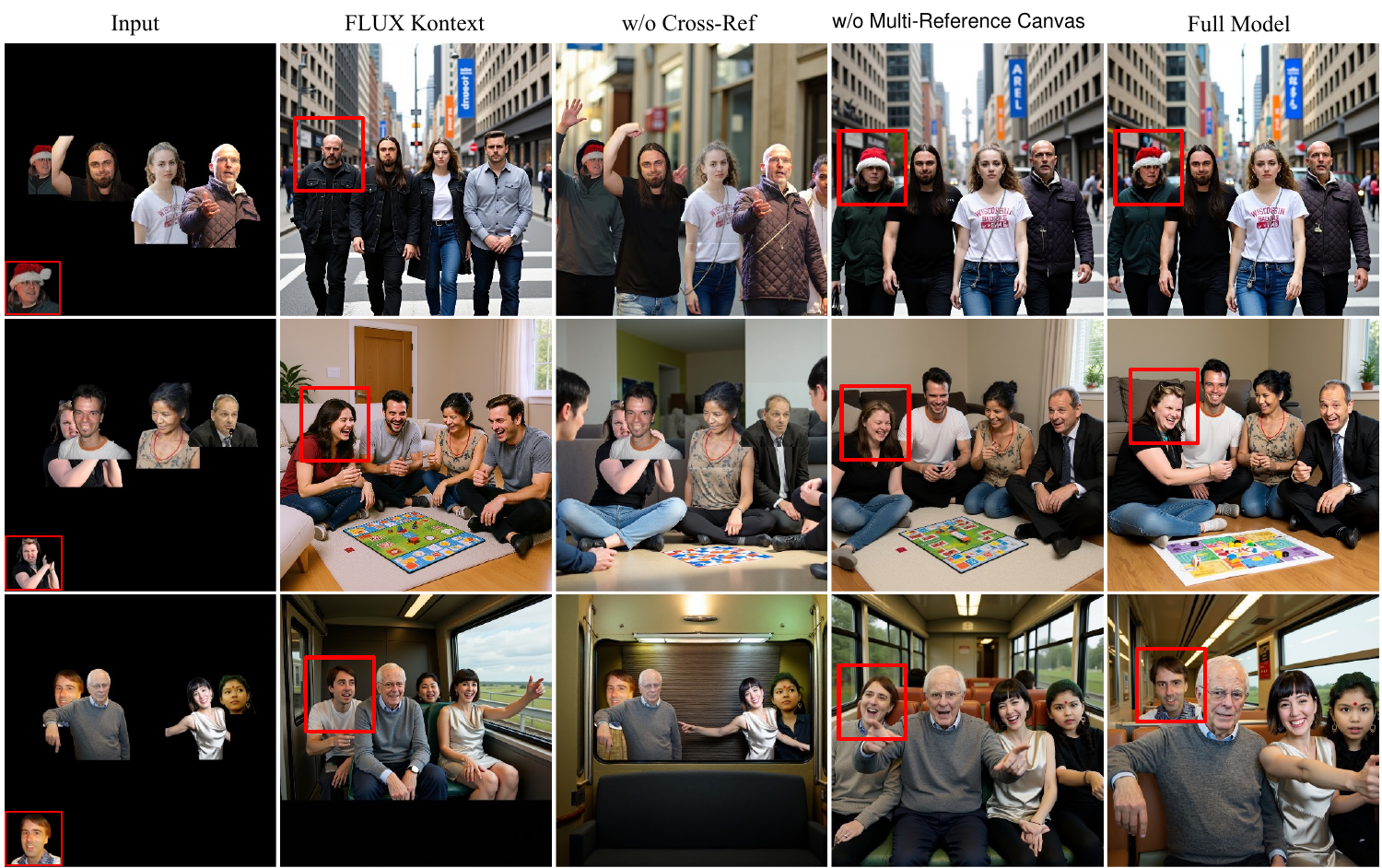}
\caption{\textbf{Ablation Study.} \rev{Cross-reference} training improves multi-subject generation by mitigating copy-paste artifacts and encouraging subjects to adapt naturally to the scene. The \rev{multi-reference} canvas addresses input occlusion (highlighted in \textcolor{red}{red boxes}): without \rev{keeping subjects on separate images}, overlapping subject details (e.g., the pom-pom on the red hat of the left woman in the $1^{\text{st}}$ row) disappear, and identity preservation deteriorates for occluded subjects.}
    \Description{Ablation grid comparing the full model with variants trained without cross-reference training and without the multi-reference canvas. The variants show copy-paste artifacts and lose details of occluded subjects, highlighted in red boxes.}
    \label{fig:ablation}
\end{figure*}

\begin{figure*}[t]
    \centering
    \newcommand{\objlbl}[2]{\makebox[#1\linewidth][c]{\scriptsize #2}}%
    \hspace*{\gidobjs\linewidth}%
    \objlbl{\lidobj}{Input}\hspace{\gidobjs\linewidth}%
    \objlbl{\lidobj}{Ours}\hspace{\gidobjb\linewidth}%
    \objlbl{\lidobj}{Input}\hspace{\gidobjs\linewidth}%
    \objlbl{\lidobj}{Ours}\hspace{\gidobjb\linewidth}%
    \objlbl{\lidobj}{Input}\hspace{\gidobjs\linewidth}%
    \objlbl{\lidobj}{Ours}\hspace*{\gidobjs\linewidth}\\[1pt]
    \includegraphics[width=\widobj\linewidth]{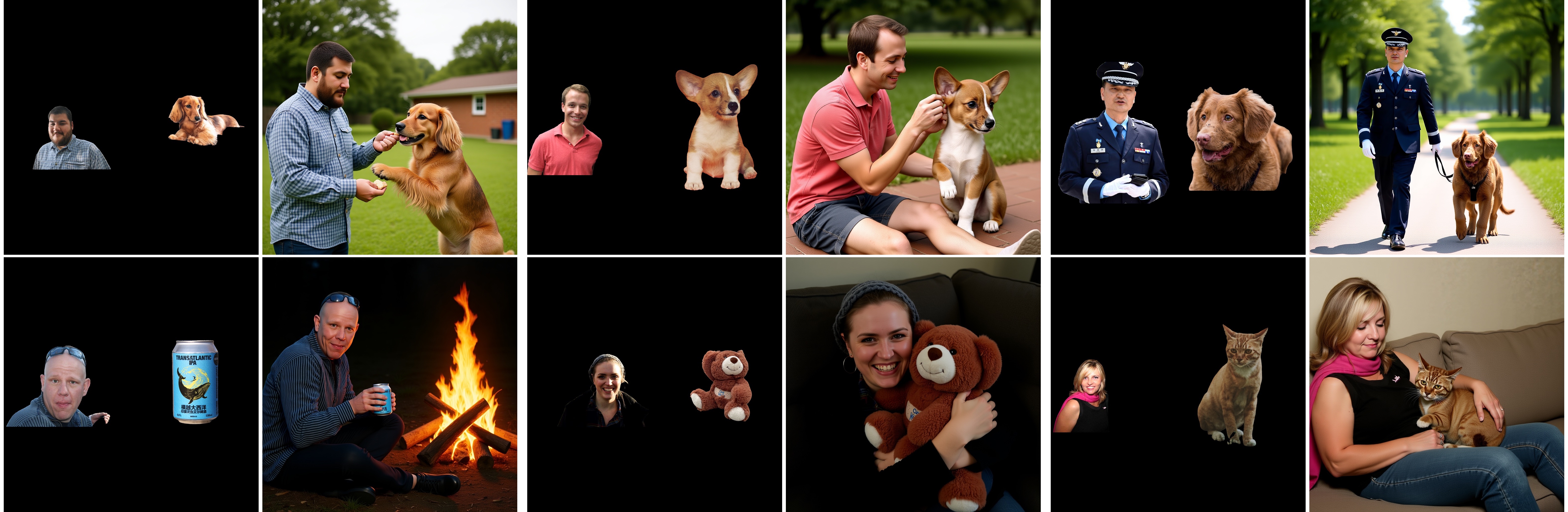}
    \caption{\textbf{Results Beyond Humans.} Although \methodname{} is trained only on human-centric data, it can accept \rev{non-human objects placed on the canvas} at inference and produce coherent interactions between people and surrounding objects.}
    \Description{Three input-output pairs in which a non-human object is placed on the canvas together with a person; the generated images show the person interacting naturally with the object.}
    \label{fig:supp:obj}
\end{figure*}

\subsection{Ablation Study and Analysis}
Our contributions primarily address interactive \rev{multi-reference} placement, occlusion handling, and training robustness through \rev{subject-wise cross-reference} training and the \rev{multi-reference canvas}.
Their effectiveness is demonstrated in \figlabel~\ref{fig:ablation}.
We include FLUX Kontext, the base model of \methodname{}, in the ablation.

\paragraph{Effect of \rev{Subject-wise Cross-Reference} Training}
Removing this component leads to copy-paste artifacts and degraded image quality, producing results that resemble naive outpainting rather than coherent multi-subject synthesis. In contrast, our full model preserves identities more naturally and avoids copying facial expressions from the inputs due to the disentanglement introduced by our training.

\begin{table}[t]
\centering
\caption{\textbf{Efficiency Analysis on the 4P Personalization Benchmark.}}
\label{tab:speed}
\resizebox{0.8\linewidth}{!}{%
\begin{tabular}{lcc}
\toprule
\textbf{Method} & \textbf{\#tokens}~$\downarrow$ & \textbf{Latency (seconds)}~$\downarrow$ \\
\midrule
\textbf{Ours} & 2200.2 & 21.2 \\
w/o \rev{Multi-Reference Canvas} & 5120 & 44.3 \\
w/o Transparent Pruning & 16384 & 91.8 \\
\bottomrule
\end{tabular}
}
\end{table}

\paragraph{Effect of \rev{Multi-Reference Canvas}}
Without the \rev{multi-reference} canvas, the model is trained on a single collage image (``Input'' column in \figlabel~\ref{fig:ablation}) without transparent latent pruning. As the number of personalized subjects increases, occlusion becomes inevitable, and such a single-canvas representation fails to preserve fine details or properly disentangle overlapping subjects.
Occluded details in the collage, e.g., the pom-pom on the red hat of the left woman in the $1^{\text{st}}$ row, are completely lost, and identity preservation for the occluded person degrades substantially, as shown in the second row.
In contrast, \methodname{} handles these scenarios robustly: \rev{keeping each subject spatially independent} preserves each reference even when \rev{subjects} overlap, preventing occlusion-induced artifacts. \rev{\cref{fig:app-reorder} further shows that the canvas layout controls the spatial arrangement of generated subjects.}

\begin{table}[t]
\centering
\caption{\textbf{\rev{Multi-reference} canvas and adapter ablations.} ``w/o \rev{Multi-Reference Canvas}'' is the collage-input baseline: the model receives a single composited RGB canvas instead of \rev{separate per-subject images}. IP-Adapter (Ours) is trained on the same data as \methodname{} under the same 2P evaluation protocol.}
\label{tab:overlap_ablation}
\setlength{\tabcolsep}{4pt}
\resizebox{\linewidth}{!}{%
\begin{tabular}{lcccccc}
\toprule
 & \multicolumn{3}{c}{\textbf{2P Personalization}} & \multicolumn{3}{c}{\textbf{2P Overlap}} \\
\cmidrule(lr){2-4}\cmidrule(lr){5-7}
\textbf{Method} & \textbf{Arc.}~$\uparrow$ & \textbf{HPS}~$\uparrow$ & \textbf{VQA}~$\uparrow$ & \textbf{Arc.}~$\uparrow$ & \textbf{HPS}~$\uparrow$ & \textbf{VQA}~$\uparrow$ \\
\midrule
\textbf{Ours} & \textbf{0.547} & \textbf{11.6} & \textbf{0.865} & \textbf{0.416} & \textbf{11.4} & \textbf{0.764} \\
w/o \rev{Multi-Reference Canvas} & 0.525 & 11.5 & 0.846 & 0.164 & 11.3 & 0.752 \\
IP-Adapter (Ours) & 0.485 & 10.2 & 0.856 & --- & --- & --- \\
\bottomrule
\end{tabular}}
\end{table}

\paragraph{Efficiency Analysis}
\cref{tab:speed} shows that transparent latent pruning reduces the average number of conditioning tokens by over $85\%$ and lowers inference latency from $91.8$s to $21.2$s on the 4P Personalization Benchmark.

\begin{figure*}[t]
    \centering
    \newcommand{\ovlbl}[2]{\makebox[#1\linewidth][c]{\scriptsize #2}}%
    \resizebox{\wovl\linewidth}{!}{%
    \ovlbl{0.09846}{Inputs}\hspace{0.00269\linewidth}%
    \ovlbl{0.19692}{\rev{Collage Baseline}}\hspace{0.00269\linewidth}%
    \ovlbl{0.19692}{Ours}\hspace{0.00462\linewidth}%
    \ovlbl{0.09846}{Inputs}\hspace{0.00269\linewidth}%
    \ovlbl{0.19692}{\rev{Collage Baseline}}\hspace{0.00269\linewidth}%
    \ovlbl{0.19692}{Ours}}\\[1pt]
    \includegraphics[width=\wovl\linewidth]{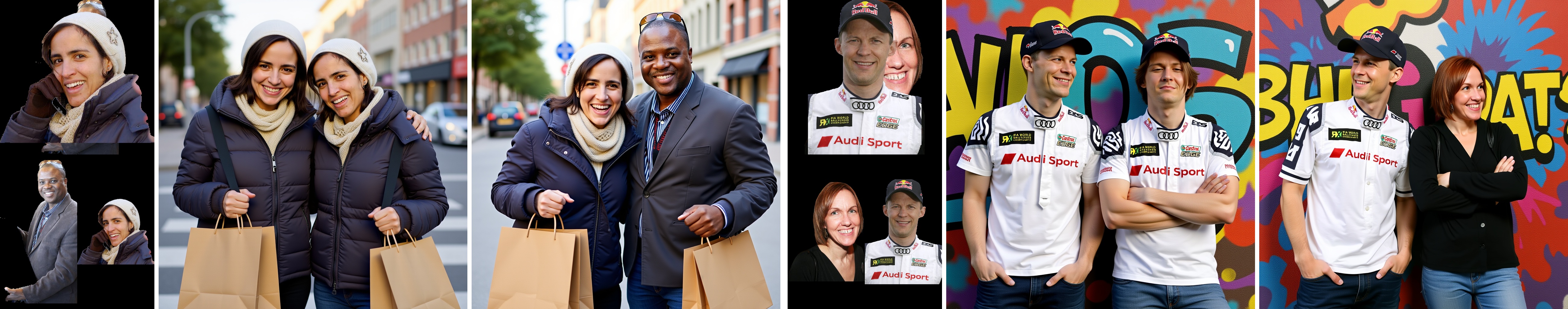}
\caption{\rev{\textbf{Qualitative Results on the 2P Overlap Benchmark.} In the Inputs column, the first row is the composited visualization of the multi-reference canvas, and the second row shows the reference inputs side by side for demonstration purposes only. The collage baseline (w/o Multi-Reference Canvas) loses the occluded reference while \methodname{} preserves both identities.}}
    \Description{Two-person overlap benchmark: for each example the inputs, the collage baseline result, and the CanvasComposer result are shown. The collage baseline loses the occluded person, CanvasComposer keeps both identities.}
    \label{fig:overlap}
\end{figure*}

\paragraph{Overlap Benchmark}
To isolate the benefit of keeping each subject \rev{spatially separate}, our 2P Overlap Benchmark centers two references in the same canvas region, so the composited RGB input occludes the background reference while \methodname{} retains \rev{both references}. \cref{tab:overlap_ablation} shows that the variant without the \rev{multi-reference} canvas is competitive on the standard 2P benchmark but loses substantial identity fidelity under overlap, while \methodname{} remains much more robust\rev{; \cref{fig:overlap} visualizes such failure modes}.

\paragraph{Adapter Baseline}
\cref{tab:overlap_ablation} also compares \methodname{} with a same-data, same-training IP-Adapter baseline on the 2P Personalization Benchmark; the \rev{multi-reference} canvas provides stronger identity preservation and image quality than adapter-style conditioning.

\paragraph{\rev{Beyond-Human Subjects}}
Although our main experiments focus on human subjects, \figlabel\ref{fig:supp:obj} shows anecdotal examples where a non-human object is \rev{placed on the canvas as an additional reference}, suggesting a possible extension of the \rev{multi-reference} representation beyond humans.

\begin{figure}[t]
    \centering
    \includegraphics[width=1.0\linewidth]{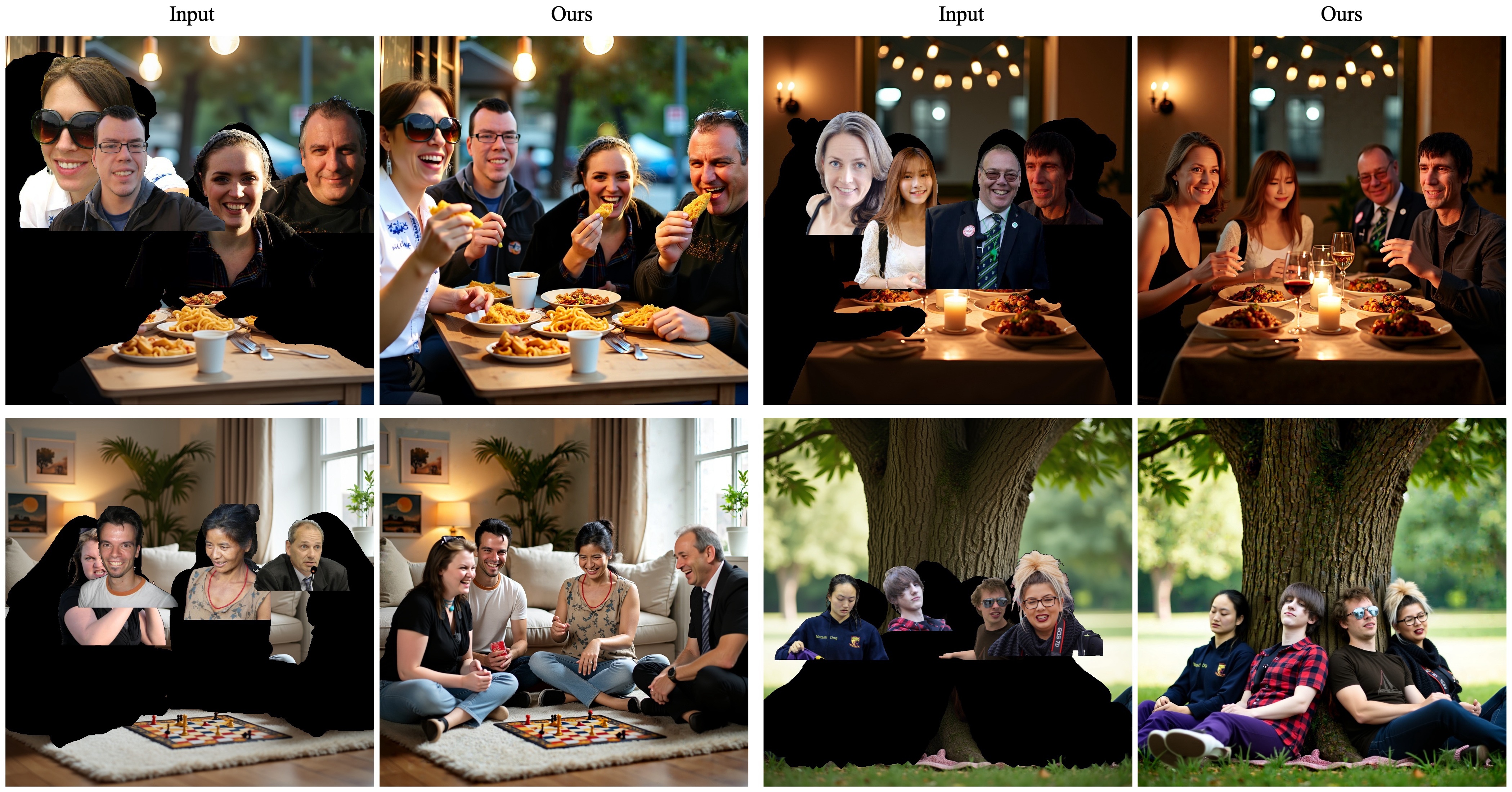}
\caption{\textbf{4P Personalization with Background.} The \rev{multi-reference} canvas can include an optional background, resulting in \rev{five separate images}: four people and one background. The inserted humans interact naturally with the background, \eg leaning against a tree trunk, under coherent lighting. Left: composited visualizations of the \rev{multi-reference} canvas.}
    \Description{Four-person group photos generated with an additional background image on the canvas; the people are inserted into the scene with consistent lighting, for example leaning against a tree trunk.}
    \label{fig:bg}
\end{figure}

\paragraph{Personalization with Background}
The \rev{multi-reference} canvas can also accept an optional background image. As shown in \figlabel\ref{fig:bg}, \methodname{} generates images where humans interact naturally with the background, \eg leaning against a tree trunk, under coherent lighting.

\section{Limitations and Future Work}\label{sec:limit}

\paragraph{Reasoning Limitations}\label{sec:limit:reasoning} The method sometimes struggles when the generated image requires detailed reasoning about how humans should interact with the background, \eg failing to seat the foreground humans naturally on chairs (see the failure example in \supp). Integrating the strong reasoning capabilities of Vision Language Models~\citep{bai2023qwen, bai2025qwen2} into personalized generation is a promising remedy.

\begin{table}[t]
\begin{minipage}[t]{0.37\linewidth}
\centering
\caption{\textbf{Beyond-4P Personalization.} Performance degrades beyond 4P.}
\label{tab:supp:beyond4p}
\resizebox{\linewidth}{!}{%
\begin{tabular}{lc}
\toprule
\textbf{Benchmark} & \textbf{ArcFace}~$\uparrow$ \\
\midrule
4P & 0.533 \\
5P & 0.206 \\
6P & 0.152 \\
\bottomrule
\end{tabular}}
\end{minipage}\hfill
\begin{minipage}[t]{0.59\linewidth}
\centering
\caption{\textbf{Base model beyond four people.} FLUX Kontext does not reliably follow prompts with more than four people.}
\label{tab:supp:base_beyond4p}
\resizebox{\linewidth}{!}{%
\begin{tabular}{cc}
\toprule
\textbf{\#People in Prompt} & \textbf{\#People in Generation} \\
\midrule
2P & 2.25 \\
4P & 4.06 \\
5P & 7.75 \\
6P & 7.75 \\
\bottomrule
\end{tabular}}
\end{minipage}
\end{table}

\paragraph{Beyond 4P Generation Limitations}
Although \methodname{} can accept a variable number of \rev{subjects}, its performance degrades beyond four people (\cref{tab:supp:beyond4p}) for two reasons.
First, \textbf{data limitations}: our $>$4P in-house samples often contain highly similar poses, expressions, or low-quality faces, so we cap the training data at four people; higher-quality, more diverse $>$4P data would likely improve robustness.
Second, \textbf{base model limitations}: FLUX Kontext itself is unreliable beyond four subjects (\cref{tab:supp:base_beyond4p}); access to its raw base weights, prior to aesthetic fine-tuning or guidance distillation (as explored for FLUX.1-Krea~\citep{blackforestlabs2025fluxkrea}), would likely enable further improvements.

\paragraph{Output Layout and Occlusion Control} \rev{The index \methodname{} uses to keep subjects separate on the input canvas carries no depth or z-ordering semantics: \methodname{} preserves reference pixels when subjects overlap in the \emph{input} canvas (\cref{tab:overlap_ablation}), while the occlusion and depth ordering of the \emph{output} follow the base model. See \seclabel\ref{sec:method_pipeline} for future directions toward such controls.}

\paragraph{Base Model Dependence} \rev{All results use a frozen FLUX Kontext base with a lightweight LoRA; performance inherits this backbone's strengths and weaknesses (\eg beyond four people, \cref{tab:supp:beyond4p,tab:supp:base_beyond4p}) and may vary on other base models.}

\paragraph{\rev{Single Flattened Output}} \rev{\methodname{} targets a single harmonized multi-human group photo and does not produce layered or transparent outputs. As a natural extension, combining our input canvas with transparent-image generation techniques~\citep{zhang2024transparent,fontanella2024rgba,li2023layerdiffusion} to produce layered RGBA outputs is compelling future work.}

\section{Conclusion}\label{sec:conclusion}
In this paper, we introduced \methodname{}, an interactive framework for multi-human personalized generation. \rev{Users can directly place and resize multiple input references on the canvas:} by representing each reference subject \rev{as its own separate image}, \methodname{} supports intuitive layout manipulation while preserving reference content under occlusion. \rev{The references are kept distinct simply by repurposing the RoPE embeddings of the pretrained base model, assigning each reference its own index.} Transparent latent pruning further reduces unnecessary conditioning tokens, and \rev{cross-reference training} mitigates copy-paste artifacts. Experiments across multi-human benchmarks show that \methodname{} improves identity preservation and generation quality over existing methods. More broadly, the \rev{multi-reference} canvas is an effective interface for adapting pretrained diffusion models to multi-human personalization. We believe \methodname{} opens several promising directions for future work.

\bibliographystyle{ACM-Reference-Format}
\bibliography{main}

\clearpage

\renewcommand{\thesection}{\Alph{section}}
\renewcommand{\thetable}{\Roman{table}}
\renewcommand{\thefigure}{\Roman{figure}}

\setcounter{section}{0}
\setcounter{table}{0}
\setcounter{figure}{0}

\section{User Study}\label{sec:supp:user}

\rev{We conduct a large-scale user study involving 20 random participants per pairwise comparison (one study per baseline), following a strictly double-anonymous procedure. To minimize evaluation bias, each participant is restricted to a single study, resulting in a total of 14,720 individual choices from 460 participants over 23 baselines across three benchmarks, as listed in the main paper and \cref{tab:supp:numbers}, with 32 prompts per benchmark.} Each participant is asked to select the better generation between two methods---our \methodname{} and a baseline---where the order of methods is randomly shuffled for every question. This pairwise evaluation yields a clearer and more interpretable quantification of user preference, reporting our win rates for each baseline on each benchmark.

\begin{table}[t]
\centering
\caption{\textbf{Supplementary quantitative comparison on the 1P Personalization Benchmark.}
\rev{All 2P/4P comparisons are reported in the main paper's quantitative table.}
\methodname{} is consistently preferred in the user study. We emphasize user preference because automatic metrics can be noisy for personalization evaluation. ArcFace favors identical pose and expression, which explains the lower 1P score for \methodname{} compared to copy-paste-style baselines.
HPSv3 and VQAScore can be noisy and less discriminative when visual quality is close.}
\label{tab:supp:numbers}
\resizebox{\linewidth}{!}{%
\begin{tabular}{lcccc}
\toprule
\textbf{Method} & \textbf{ArcFace}~$\uparrow$ & \textbf{HPSv3}~$\uparrow$ & \textbf{VQAScore}~$\uparrow$ & \textbf{Ours Win Rate (\%)}~$\uparrow$ \\
\midrule
\multicolumn{5}{c}{\textbf{1P Personalization} (\textit{Single-Subject Personalization Baselines})} \\
IP-Adapter~\cite{ye2023ip-adapter} & 0.453 & 9.88 & 0.790 & 61.9 \\
PuLID-FLUX~\cite{PuLID} & 0.639 & 11.5 & 0.859 & 62.5 \\
InfiniteYou~\cite{InfiniteYou} & 0.528 & 13.2 & 0.853 & 68.3 \\
\multicolumn{5}{c}{\textbf{1P Personalization}  (\textit{Multi-Subject Personalization Baselines})} \\
UniPortrait~\cite{he2024uniportrait} & 0.452 & 13.6 & 0.876 & 63.8 \\
StoryMaker~\cite{StoryMaker} & 0.589 & 5.49 & 0.619 & 76.6 \\
ID-Patch~\cite{ID-Patch} & 0.124 & 10.6 & 0.911 & 85.0 \\
UNO~\cite{UNO} & 0.161 & 12.3 & 0.874 & 77.5 \\
DreamO~\cite{DreamO} & 0.694 & 13.4 & 0.852 & 57.0 \\
OmniGen2~\cite{wu2025omnigen2} & 0.305 & 11.8 & 0.883 & 58.4 \\
\textbf{Ours} & 0.487 & 12.5 & 0.893 & -- \\
\bottomrule
\end{tabular}}
\end{table}

\begin{table*}[p]
\centering
\setlength{\tabcolsep}{2pt} 
\renewcommand{\arraystretch}{0.1} 
\renewcommand{\thetable}{\thefigure}
\captionsetup{type=figure}
\begin{tabularx}{0.9\linewidth}{ *{5}{>{\centering\arraybackslash}X} }
\scriptsize \textbf{IP-Adapter}~\cite{ye2023ip-adapter} & \scriptsize \textbf{PuLID}~\cite{PuLID} & \scriptsize \textbf{InfiniteYou}~\cite{InfiniteYou} & \scriptsize \textbf{Ours} & \scriptsize \textbf{Inputs}\\[3pt]

\multicolumn{5}{c}{\hspace*{-\tabcolsep}{\includegraphics[width=0.92\linewidth]{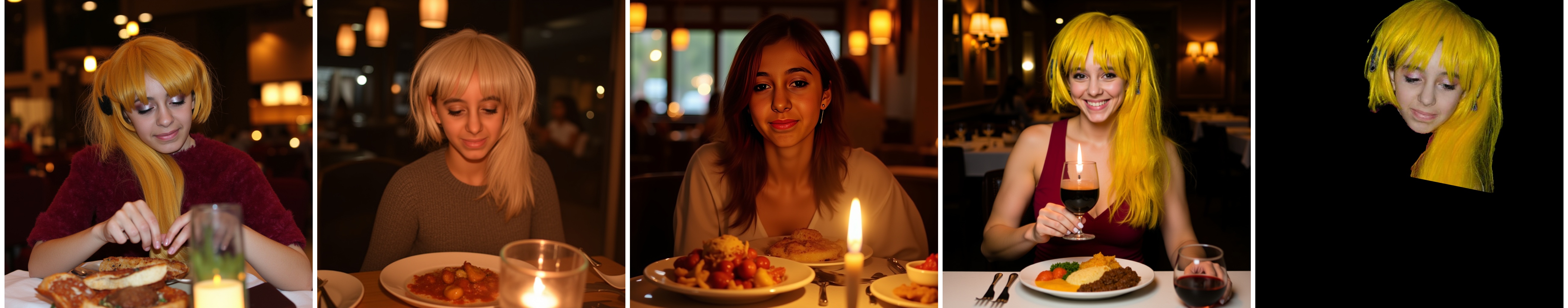}}} \\
\multicolumn{5}{c}{\hspace*{-\tabcolsep}{\includegraphics[width=0.92\linewidth]{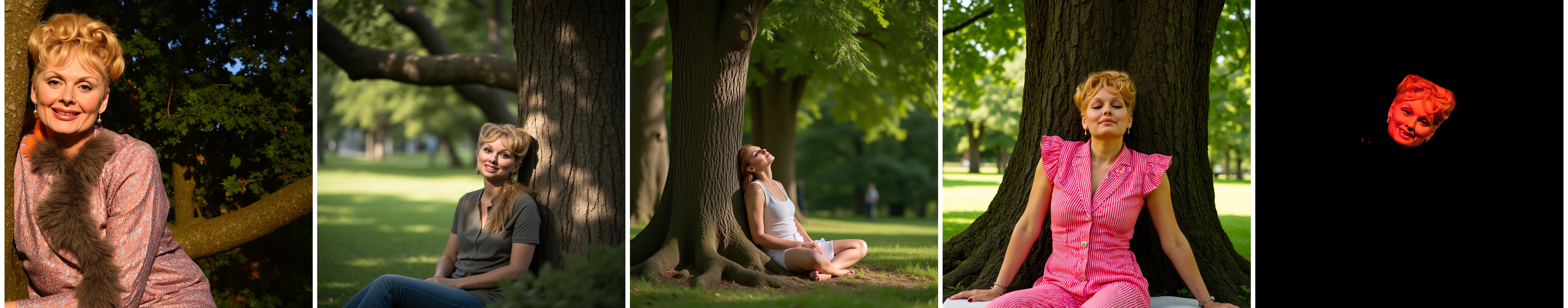}}} \\
\multicolumn{5}{c}{\hspace*{-\tabcolsep}{\includegraphics[width=0.92\linewidth]{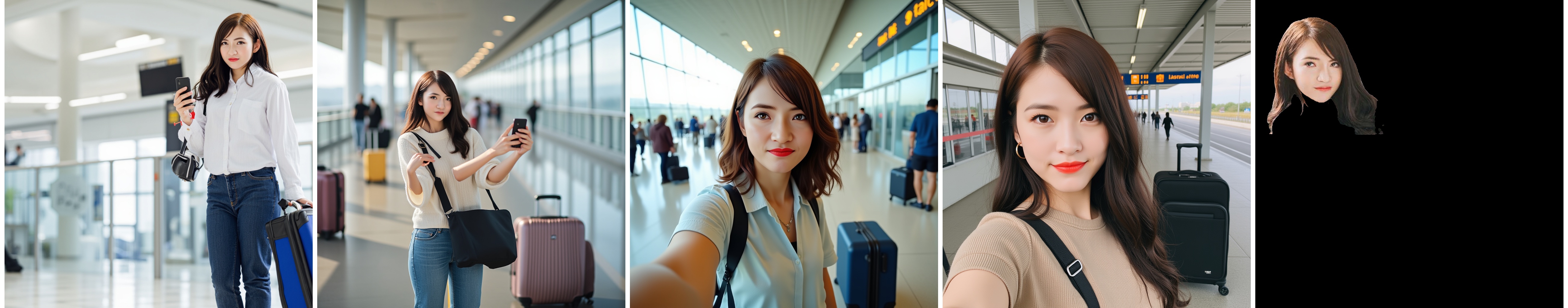}}} \\
\multicolumn{5}{c}{\hspace*{-\tabcolsep}{\includegraphics[width=0.92\linewidth]{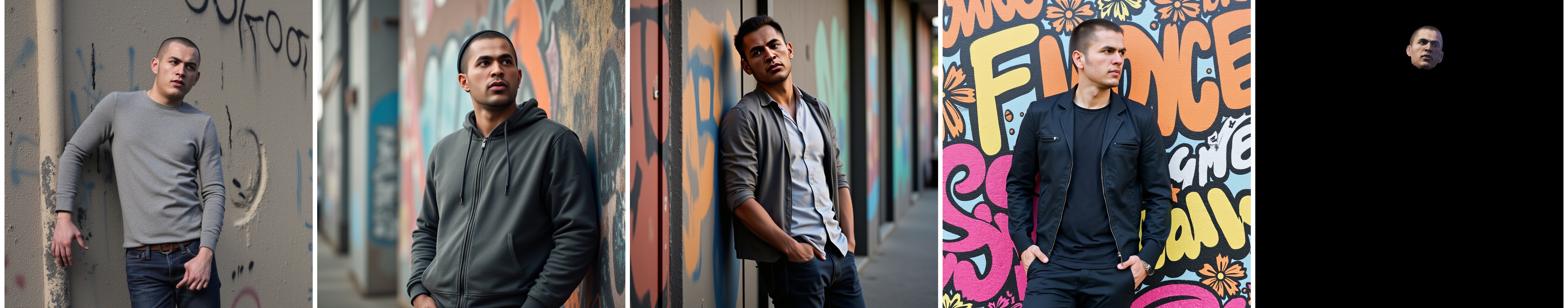}}} \\
\multicolumn{5}{c}{\hspace*{-\tabcolsep}{\includegraphics[width=0.92\linewidth]{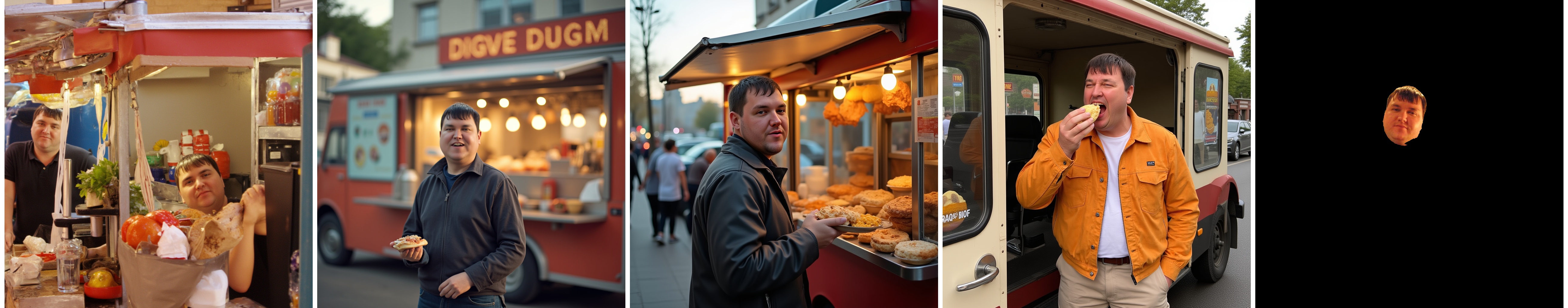}}} \\
\multicolumn{5}{c}{\hspace*{-\tabcolsep}{\includegraphics[width=0.92\linewidth]{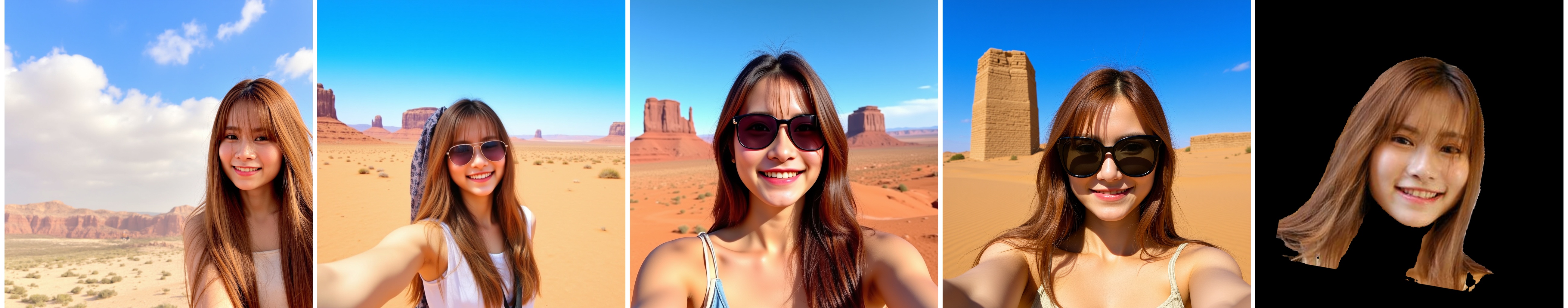}}} \\
\end{tabularx}
\caption{\textbf{Qualitative Comparison in Single-Person (1P) Personalization.}
    State-of-the-art 1P personalization approaches often preserve the reference face with limited flexibility, resulting in copy-paste effects. In contrast, \methodname{} generates realistic outputs that are faithful to both the human identity and the text prompt. Notably, our method captures diverse expressions (e.g., \emph{smiling}, 1\textsuperscript{st} row), handles challenging states such as \emph{relaxing} and \emph{closed eyes} (2\textsuperscript{nd} row), and supports activities like \emph{eating} (5\textsuperscript{th} row), which require complex body poses or expressive facial gestures. All baselines are based on FLUX.1 dev~\cite{FLUX} and use the cropped head as input. \methodname{} uses the same cropped head \rev{as the sole reference image}.}
\label{fig:supp:1p}
\end{table*}

\section{Additional Results}
Since \methodname{} is designed for multi-subject personalization, our main paper focuses on comparisons against multi-subject personalization baselines. Here, we additionally extend the evaluation to include image-editing methods for the 4P Personalization Benchmark and single-subject personalization approaches for the 1P Personalization benchmark.

\begin{table*}[p]
\centering
\setlength{\tabcolsep}{2pt} 
\renewcommand{\arraystretch}{0.1} 
\renewcommand{\thetable}{\thefigure}
\captionsetup{type=figure}
\begin{tabularx}{0.95\linewidth}{ *{6}{>{\centering\arraybackslash}X} }

\scriptsize FLUX Kontext~\cite{fluxkontext} & \scriptsize {Overlay Kontext~\cite{ilkerzgi2025overlay}} & \scriptsize {Qwen-Image-Edit}~\cite{qwen-image} & \scriptsize {Nano-Banana}~\cite{comanici2025gemini} & \scriptsize \textbf{Ours} & \scriptsize \textbf{Inputs}\\[-6pt]

\multicolumn{6}{c}{\hspace*{-\tabcolsep}{\includegraphics[width=0.95\linewidth]{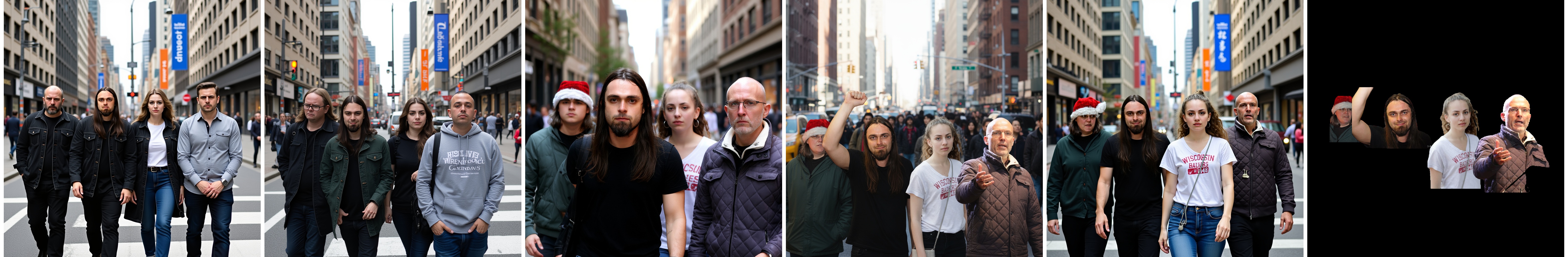}}} \\
\multicolumn{6}{c}{\hspace*{-\tabcolsep}{\includegraphics[width=0.95\linewidth]{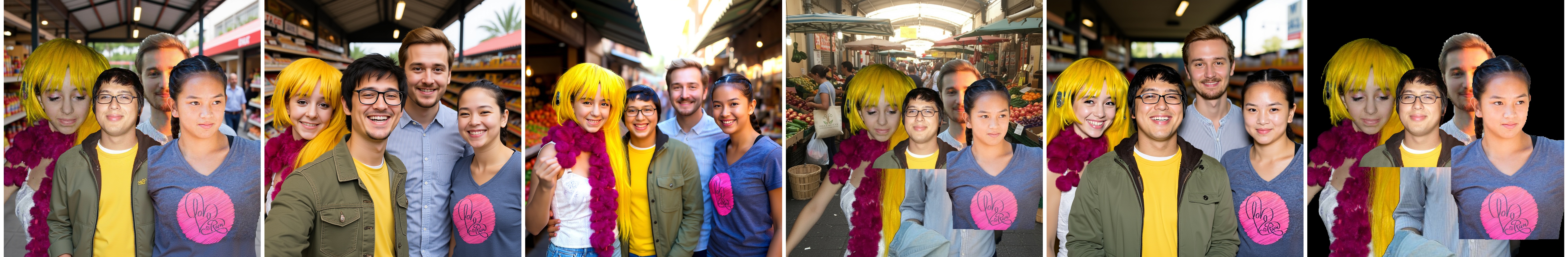}}} \\
\multicolumn{6}{c}{\hspace*{-\tabcolsep}{\includegraphics[width=0.95\linewidth]{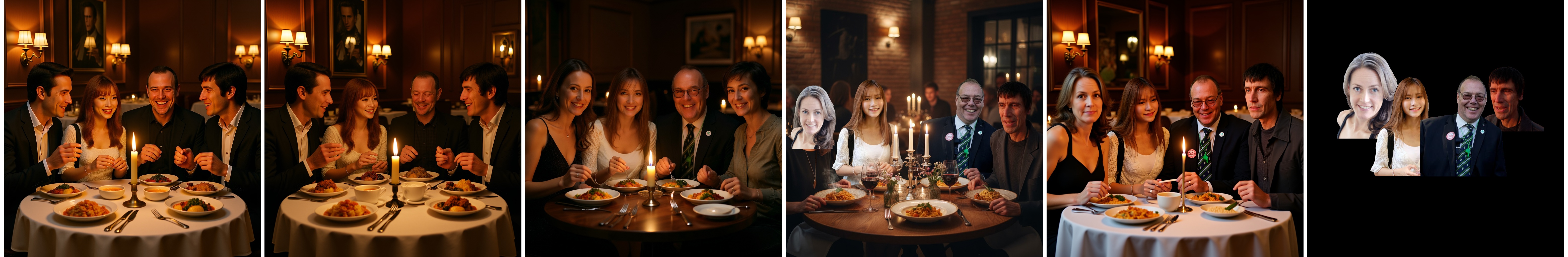}}} \\
\multicolumn{6}{c}{\hspace*{-\tabcolsep}{\includegraphics[width=0.95\linewidth]{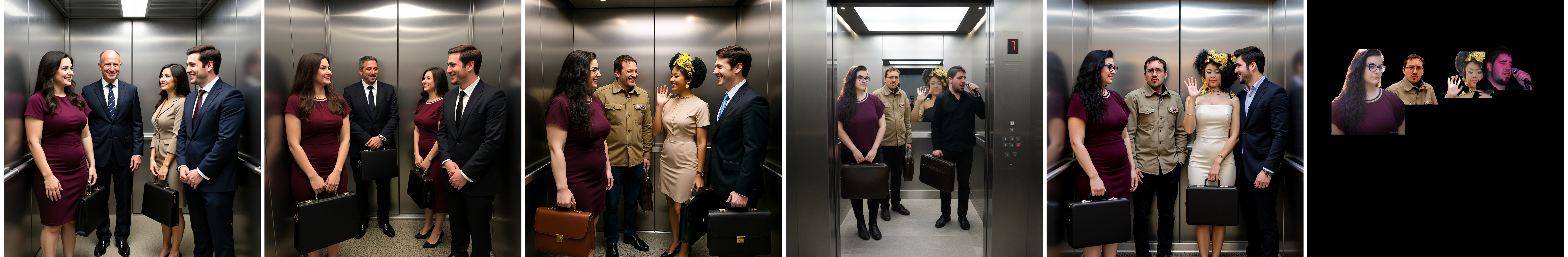}}} \\
\multicolumn{6}{c}{\hspace*{-\tabcolsep}{\includegraphics[width=0.95\linewidth]{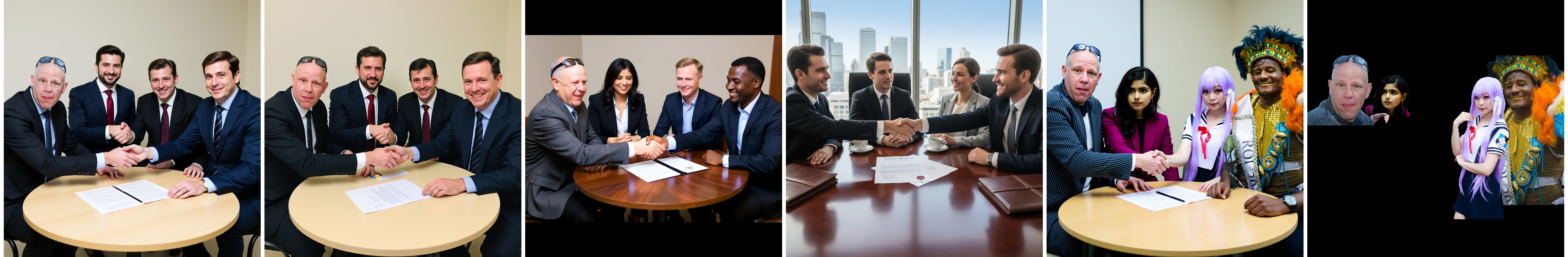}}} \\
\multicolumn{6}{c}{\hspace*{-\tabcolsep}{\includegraphics[width=0.95\linewidth]{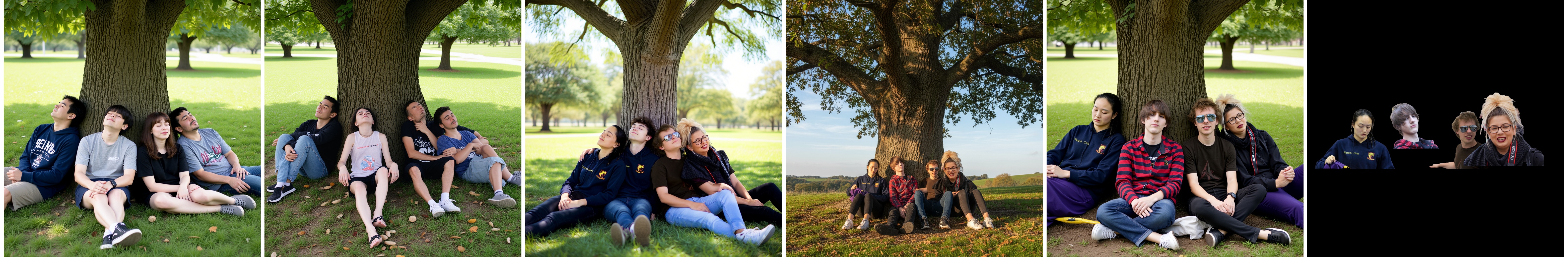}}} \\
\multicolumn{6}{c}{\hspace*{-\tabcolsep}{\includegraphics[width=0.95\linewidth]{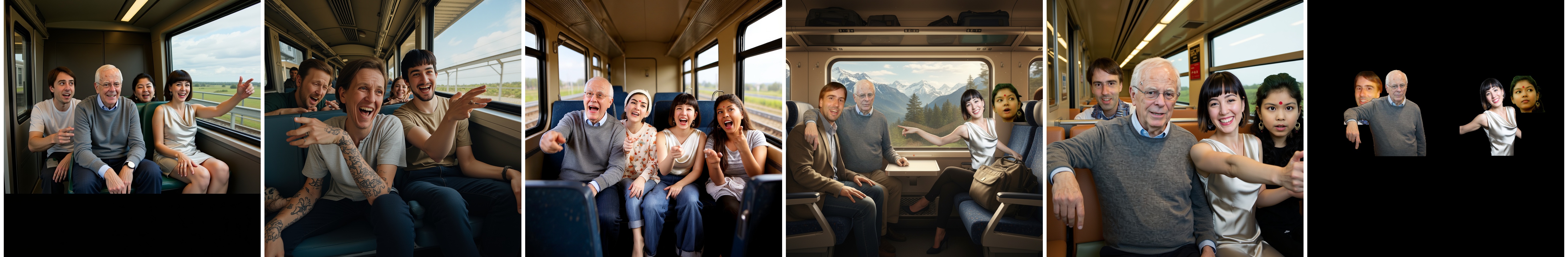}}} \\
\end{tabularx}
\caption{\textbf{Qualitative Comparison to Image-Editing Baselines in Four-person (4P) Personalization.} State-of-the-art image-editing methods frequently distort or omit subjects, or produce unnatural copy-paste artifacts, because they are not designed for multi-subject personalization. In contrast, \methodname{} consistently generates high-fidelity results while preserving identities in challenging multi-person scenes. Although \methodname{} builds on FLUX Kontext~\cite{fluxkontext}, it achieves stronger performance due to our \rev{multi-reference} canvas design and \rev{cross-reference} training.}
\label{fig:supp:4p}
\end{table*}

\subsection{Comparison with Image Editing Pipelines in 4P Personalization}\label{sec:supp:4p}
Four-person (4P) personalization is a particularly challenging setting; the main paper reports the quantitative comparisons. Here, we additionally provide qualitative comparisons against recent image-editing methods that have emerged as competitive baselines for multi-subject personalization, including FLUX Kontext~\cite{fluxkontext}, Overlay Kontext~\cite{ilkerzgi2025overlay} (Place-It LoRA), Qwen-Image-Edit~\cite{qwen-image}, and Nano-Banana~\cite{comanici2025gemini}.
Since these methods do not support the \rev{multi-reference} canvas, we provide them with collaged inputs. All models, including \methodname{}, are evaluated using identical prompts. For Overlay Kontext, we follow official guidelines and prepend the trigger phrase \texttt{``Place it.''} to each prompt.

Occlusions naturally arise as the number of personalized subjects increases, and \methodname{} handles these cases through its \rev{multi-reference} canvas design.
By contrast, baseline approaches often fail under such conditions, as shown in \cref{fig:supp:4p}.

Quantitative results for all image-editing baselines are reported in the main paper's quantitative table.
While \methodname{} adopts FLUX Kontext~\cite{fluxkontext} as its base model, the \rev{multi-reference-canvas} representation and \rev{cross-reference} training improve performance over FLUX Kontext, as shown in \cref{fig:supp:4p}.

\subsection{1P Personalization Benchmark across Single-Subject Personalization Methods}\label{sec:supp:1p}
In the main paper, we focused on multi-subject scenarios. Here, we additionally evaluate \methodname{} on the 1P (single-person) personalization benchmark (\cref{fig:supp:1p}, \cref{tab:supp:numbers}).
We compare against leading single-subject methods built upon FLUX.1 dev~\citep{FLUX}, including IP-Adapter~\citep{ye2023ip-adapter}, PuLID~\citep{PuLID}, and InfiniteYou~\citep{InfiniteYou}. 

As shown in \figlabel~\ref{fig:supp:1p}, competing methods frequently inject the reference identity directly with limited variation in pose and expression, limiting their ability to accurately follow diverse prompts. In contrast, \methodname{} produces coherent and natural generations that faithfully adhere to text prompts while maintaining identity fidelity and diverse facial expressions.

Quantitatively, \tablabel~\ref{tab:supp:numbers} shows that \methodname{} attains the highest prompt adherence among all baselines that preserve identity (VQAScore $0.893$; the higher $0.911$ of ID-Patch~\cite{ID-Patch} comes with an ArcFace of only $0.124$, \ie the reference identity is not preserved), and is preferred by users over every single-subject and multi-subject baseline, with win rates ranging from $57.0\%$ to $85.0\%$.

\begin{figure}
    \centering
    \includegraphics[width=0.7\linewidth]{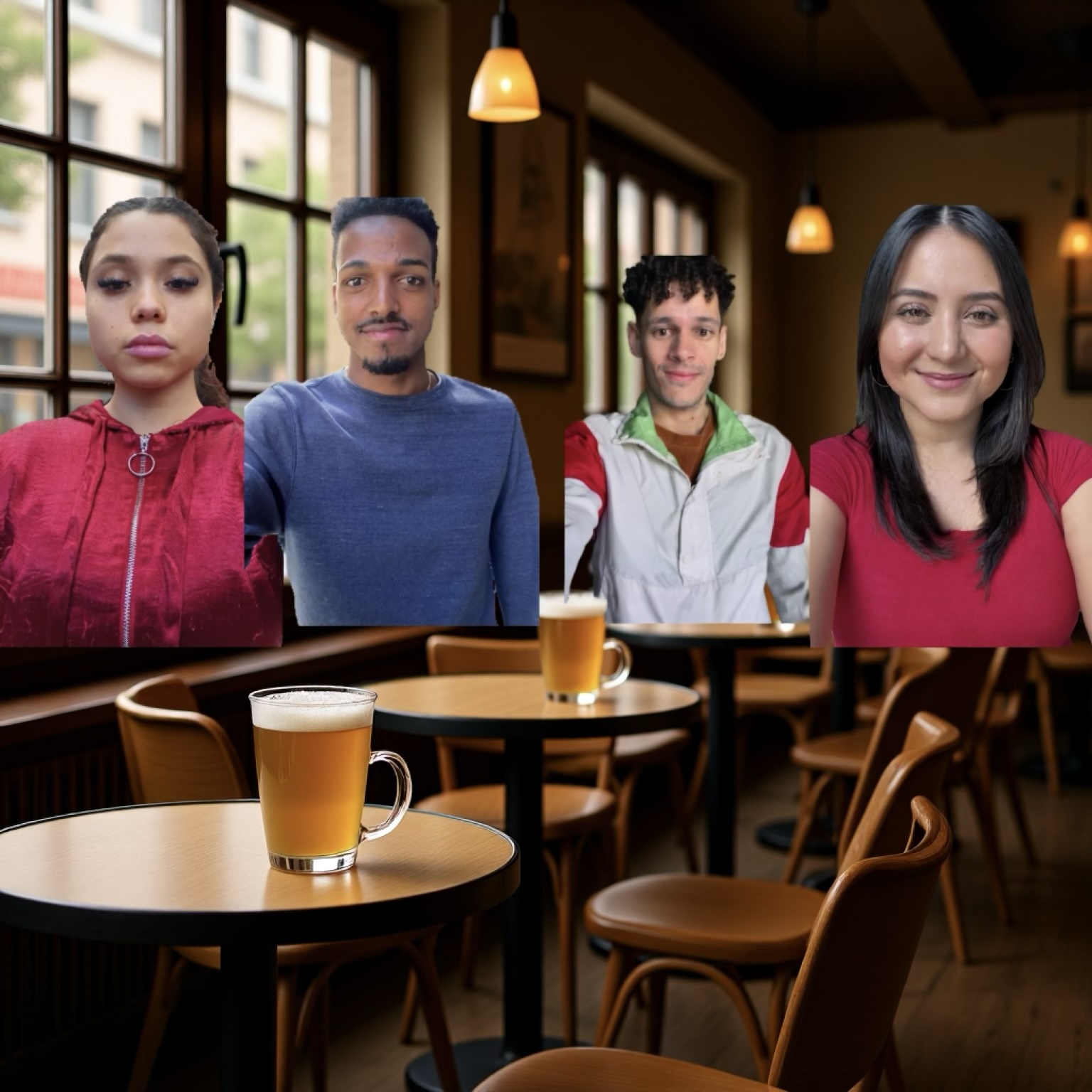}
    \caption{
    \textbf{Failure Sample in Complex Reasoning Scenarios.}
    \methodname{} struggles when strong interaction reasoning is required: the subjects fail to sit naturally on the chairs, yielding copy-paste-like outputs that closely resemble the inputs.
    }
    \Description{Failure case: four subjects asked to sit around a coffee table. The generated people stand stiffly instead of sitting naturally, closely resembling the reference cut-outs.}
    \label{fig:limitation}
\end{figure}

\paragraph{Failure Case} \rev{\cref{fig:limitation} shows a failure example in a complex reasoning scenario, complementing the Limitations section of the main paper.}

\section{Additional Ablations}\label{sec:supp:ablation}

\begin{table}[t]
\centering
\caption{\textbf{LoRA rank ablation on the 4P Personalization Benchmark.} Rank 512 provides the strongest identity preservation while maintaining competitive image quality and prompt alignment.}
\label{tab:supp:rank_ablation}
\resizebox{0.75\linewidth}{!}{%
\begin{tabular}{lccc}
\toprule
\textbf{Rank Size} & \textbf{ArcFace}~$\uparrow$ & \textbf{HPSv3}~$\uparrow$ & \textbf{VQAScore}~$\uparrow$ \\
\midrule
Ours (rank 512) & \textbf{0.533} & 12.5 & 0.84 \\
rank 32 & 0.521 & 12.0 & 0.85 \\
rank 128 & 0.524 & 12.3 & 0.84 \\
rank 1024 & 0.521 & \textbf{12.6} & \textbf{0.86} \\
\bottomrule
\end{tabular}}
\end{table}

\paragraph{LoRA Rank}
We ablate the LoRA rank on the 4P Personalization Benchmark in \cref{tab:supp:rank_ablation}. Rank 512 gives the strongest ArcFace score while preserving competitive HPSv3 and VQAScore, so we use it as the default setting.

\section{Full 3D RoPE Formulation}\label{sec:supp:rope}
{\revblock
\methodname{} reuses the factorized 3D Rotary Position Embedding (RoPE)~\cite{su2024roformer} of FLUX Kontext~\cite{fluxkontext} without modification; here we state the full formulation for completeness.
Each token carries a position $\text{pos} = (j, x, y)$, where $j$ \rev{identifies which canvas element the token belongs to} and $(x, y)$ are latent-grid coordinates. The per-head channel dimension $d$ is partitioned into three disjoint groups of sizes $d_j$, $d_x$, and $d_y$ (with $d_j + d_x + d_y = d$), assigned to the three axes following the base model's configuration. For an axis with coordinate $p$ and group size $d_a$, RoPE rotates each consecutive channel pair $(2k, 2k{+}1)$, $k \in \{0, \ldots, d_a/2 - 1\}$, by the angle $p\,\theta^{-2k/d_a}$ with base $\theta = 10^4$:
\begin{equation}
    R_{a}(p) = \bigoplus_{k=0}^{d_a/2-1}
    \begin{pmatrix}
    \cos p\theta_k & -\sin p\theta_k \\
    \sin p\theta_k & \cos p\theta_k
    \end{pmatrix},
    \quad \theta_k = \theta^{-2k/d_a},
\end{equation}
where $\bigoplus$ denotes a block-diagonal composition. The full 3D rotation applied to the query and key vectors of a token is the block-diagonal composition over the three axes,
\begin{equation}
    R(\text{pos}) = R_{j}(j) \oplus R_{x}(x) \oplus R_{y}(y),
\end{equation}
so attention logits depend only on relative offsets along each axis. \methodname{} keeps $d_j$, $d_x$, $d_y$, and $\theta$ identical to the base model and repurposes only \rev{the element-identity axis}: $j = 0$ is reserved for the noisy latent tokens, and $j \ge 1$ indexes \rev{the canvas's input elements}.
}

\section{Benchmark Details}\label{sec:supp:benchmark}
\subsection{Baselines}
In terms of prompts, for OmniGen2~\citep{wu2025omnigen2}, which requires in-context instructions, we follow its required prompting format: 
\texttt{"The first person is image 1 and the second person is image 2.\{prompt\}."} for 2P generation, and 
\texttt{"The person is in image 1.\{prompt\}."} for 1P generation. All other methods, including \methodname{}, use the \texttt{"\{prompt\}."} directly. 

In terms of inputs, 
for \methodname{}, we use the \rev{multi-reference canvas whose composited visualization is shown in} the ``Inputs'' column \rev{of} each figure\rev{.}
For image editing pipelines~\cite{comanici2025gemini, fluxkontext, ilkerzgi2025overlay,qwen-image}, we use the ``Inputs'' column directly as input as they do not support \rev{the multi-reference canvas}.
For all previous personalization methods, we use the individual input images as inputs to each method as required.

\subsection{Automated Benchmarking Pipeline}
The \rev{multi-reference} canvas is primarily designed for interactive personalization. To enable reproducible benchmarking without human intervention, we develop an automated canvas-creation pipeline. For each prompt, we first generate a prior image using FLUX.1 dev~\citep{FLUX}. We then apply face detection to identify people in the prior image, followed by detecting the face regions in all input images corresponding to the personalized subjects. Each input image is segmented using \rev{our in-house YOLO11-style~\citep{yolo11} human instance segmenter} and assigned to the matching detected subject, forming a \rev{multi-reference} canvas that serves as the input to our model. A \rev{composited visualization} is created from this \rev{multi-reference} canvas and used \rev{as the input} for all other image editing baselines. This \rev{composited image} corresponds to the ``Inputs'' column in each figure. This automated pipeline is applied to both 4P and 2P personalization. For 1P personalization, where only a cropped head is used as the reference, we use face landmarks to create \rev{the corresponding single-subject RGBA image}.

\section{Ethics and Broader Impact}\label{sec:supp:ethics}
{\revblock
\paragraph{Image Rights} All subject images used for evaluation come from public datasets (FFHQ-in-the-wild~\cite{StyleGAN}) or are otherwise licensed for research use. Our in-house training data cannot be released due to internal data policies; following~\cite{Omni-ID}, a comparable dataset can be curated from publicly available videos, as noted in the main paper.

\paragraph{Potential Misuse} Like other personalized human-generation methods, \methodname{} could be misused to synthesize images of real people without their consent, e.g., for impersonation or misinformation. We intend the method for creative and consented applications only, and we encourage the use of reference images obtained with consent together with provenance signals such as watermarking or content credentials.

\paragraph{Demographic Fairness} Generation quality and identity preservation may vary across demographic groups because of biases inherited from the training data and the frozen base model. We recommend evaluating and mitigating such disparities before deployment in downstream products.
}

\end{document}